%% file: root.tex
\documentclass[letterpaper, 10 pt, conference]{ieeeconf}  %

\usepackage{graphicx}
\usepackage{amssymb}
\usepackage{amsmath}
\usepackage{commath}
\usepackage{mathtools}

\usepackage{hyperref}
\usepackage{import}
\usepackage{paralist}
\usepackage{gensymb}
\usepackage{dsfont}
\usepackage{siunitx}
\usepackage[bottom]{footmisc}
\usepackage{framed}
\usepackage{balance} 
\usepackage{color, colortbl}
\usepackage[dvipsnames]{xcolor}
\usepackage[skins]{tcolorbox}
\usepackage{url}
\usepackage{cleveref}
\usepackage{microtype}
\usepackage{tikzscale} %
\usepackage{pgfplots}

\usepackage{todonotes}

\usepackage{textcomp}
\usepackage{gensymb}
\usepackage[caption=false]{subfig}

\pgfplotsset{compat=1.17}

\newsavebox{\measurebox}

\usepackage{amsmath,mathtools}

\newtcolorbox{myframe_bak}[2][]{%
  enhanced,colback=white,colframe=black,coltitle=black,
  sharp corners,boxrule=0.4pt,left=0pt,right=0pt,top=0pt,bottom=0pt,
  attach boxed title to top left={yshift=-0.3\baselineskip-0.4pt,xshift=2mm},
  boxed title style={tile,size=minimal,left=1mm,right=1mm,
  colback=white,before upper=\strut},
  title=#2,#1
}

\newtcolorbox{myframe}[2][]{%
  enhanced,colback=white,colframe=black,coltitle=black!75!white,
  left=0pt,right=0pt,top=0pt,bottom=0pt,
  attach boxed title to top left={yshift=-0.4\baselineskip-0.4pt,xshift=1.5mm},
  boxed title style={tile,size=minimal,left=1mm,right=1mm,
  colback=white,before upper=\strut},
  title=#2,#1
}

\IEEEoverridecommandlockouts                              %

\title{\LARGE \bf
Whole-Body Trajectory Optimization for Robot Multimodal Locomotion
}

\author{Giuseppe L'Erario$^{1,2}$, Gabriele Nava$^{1}$, Giulio Romualdi$^{1}$, \\ Fabio Bergonti$^{1,2}$, Valentino Razza$^{1}$, Stefano Dafarra$^{1}$ and Daniele Pucci$^{1,2}$%
\thanks{$^{1}$ Artificial and Mechanical Intelligence, Istituto Italiano di Tecnologia, Genoa, Italy
        {\tt\footnotesize name.surname@iit.it}}%
\thanks{$^{2}$ University of Manchester, Manchester, UK}
\thanks{This work is supported by Honda R\&D Co.}
}

\begin{document}

\maketitle
\thispagestyle{empty}
\pagestyle{empty}

\begin{abstract}

The general problem of planning feasible trajectories for multimodal robots is still an open challenge. This paper presents a whole-body trajectory optimisation approach that addresses this challenge by combining methods and tools developed for aerial and legged robots. First, robot models that enable the presented whole-body trajectory optimisation framework are presented. The key model is the so-called robot centroidal  momentum, the dynamics of which is directly related to the models of the robot actuation for aerial and terrestrial locomotion. Then, the paper presents how these models can be employed in an optimal control problem to generate either terrestrial or aerial locomotion trajectories with a unified approach. The optimisation problem considers robot kinematics, momentum, thrust forces and their bounds. The overall approach is validated using the multimodal robot iRonCub, a flying humanoid robot that expresses a degree of terrestrial and aerial locomotion. To solve the associated optimal trajectory generation problem, we employ ADAM, a custom-made open-source library that implements a collection of algorithms for calculating rigid-body dynamics using CasADi. 
\end{abstract}

\input{tex/custom_commands}
\input{tex/introduction}

\input{tex/background}

\input{tex/method}
\input{tex/results}

\input{tex/conclusions}

\bibliography{references}
\bibliographystyle{IEEEtran}

\end{document}

%% file: tex/custom_commands.tex
\newcommand{\asFrame}[1]{\mathcal{#1}}

\newcommand{\R}{\mathbb{R}}

\newcommand{\com}{x_\text{CoM}}

%% file: tex/introduction.tex
\section{Introduction}

Monomodal locomotion systems, such as legged robots or quadrotors, often struggle when dealing with diverse scenarios. For instance, legged robots may traverse cluttered environments but often strive when covering large distances. Quadrotors and drones, instead, may cover larger distances at the cost of large energy expenditures when performing quasi-static tasks where legged systems are more efficient. 
Robots with multimodal locomotion abilities can overcome these difficulties that monomodal systems have, thus offering potential solutions for several real applications. Their possibility to use either terrestrial or aerial locomotion renders multimodal robots more effective when the surrounding environment and the underlying terrain are unstructured and difficult to predict. 
Flying exapods~\cite{Pitonyak2017}, insect biobots~\cite{bozkurt2009aerial}, advanced quadrotors~\cite{Kalantari2013}, and flying humanoid robots~\cite{Nava2018PositionRobot} are only a few examples of multimodal robots that may have a concrete impact in the near future~\cite{daler2015bioinspired, kondak2013unmanned, kim2021bipedal}. 
One of the main questions that arise when controlling multimodal robots, however, is the automatic transition between aerial and terrestrial locomotion, which is supposed to depend upon the robot's environment and desired motion. This paper takes a step in this direction by proposing whole-body trajectory optimisation methods that generate multimodal locomotion patterns depending on the robot's desired motion and surrounding environment.

Planning trajectories for multimodal robots can leverage tools and methods developed for flying and legged systems.
The literature on flying vehicles is large and diverse~\cite{xilun2019review, Ruggiero2018AerialReview}. Most of the research is grounded on the assumption that the aerial system is a \textit{floating base} rigid-body powered by body-fixed thrust forces and that the aerodynamic forces are considered negligible. Common strategies to generate open-loop trajectories are based on RRT algorithms~\cite{lavalle1998rapidly, lavalle2001randomized} as well on optimisation-based approaches~\cite{richter2016polynomial}. The output of the trajectory generation layer is then fed to a controller whose aim is to stabilize the planned trajectory. A widely used control approach is the so-called \textit{vectored-thrust} paradigm~\cite{naldi2016robust, lee2010geometric}: having modelled the quadrotor as a rigid body, we can exploit the coupling between the robot position and attitude and use the angular velocity to align the external forces to the planned trajectory.

\begin{figure}[tpb]
\centering
\includegraphics[width=.4\linewidth]{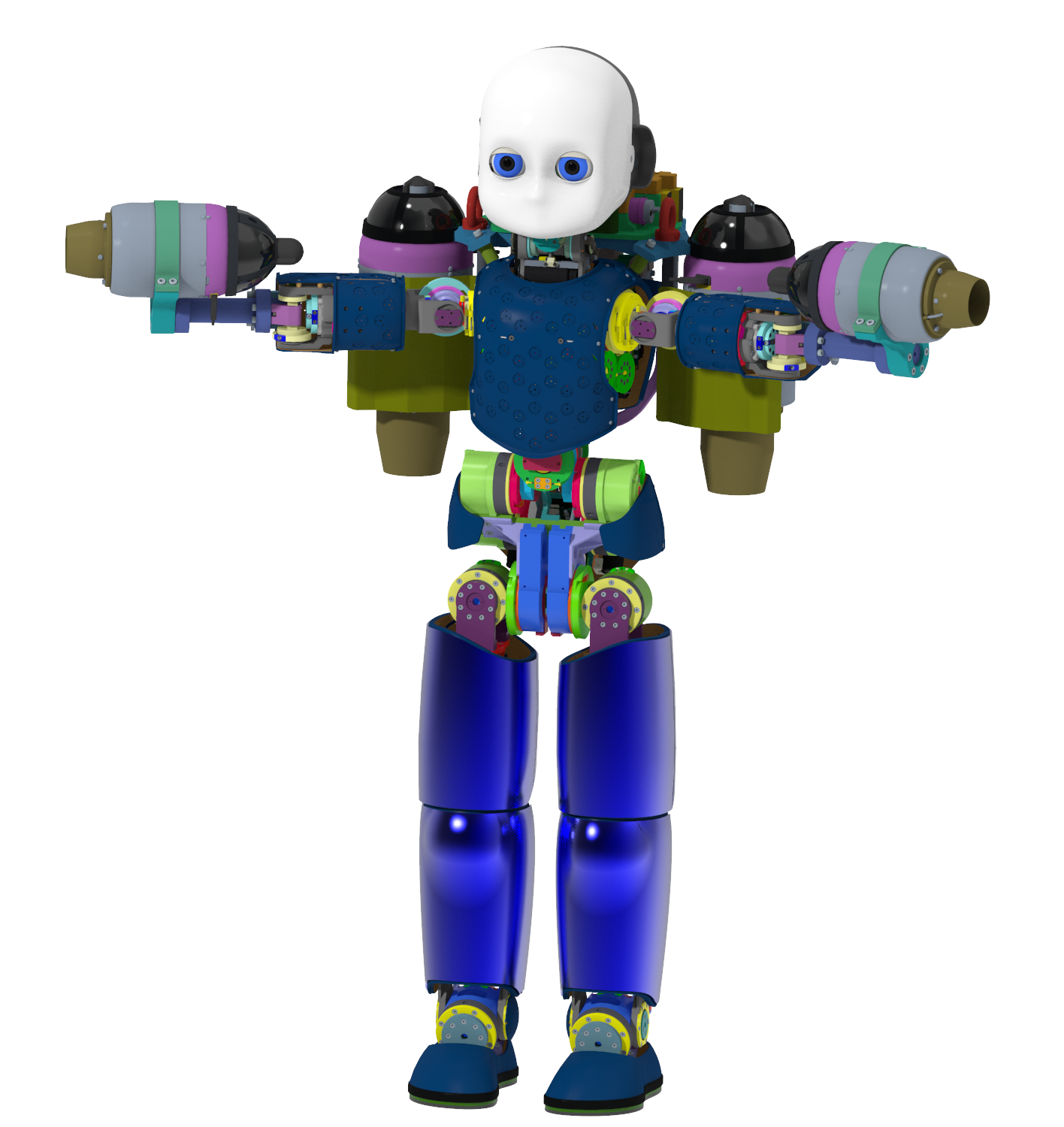}
\includegraphics[width=.4\linewidth]{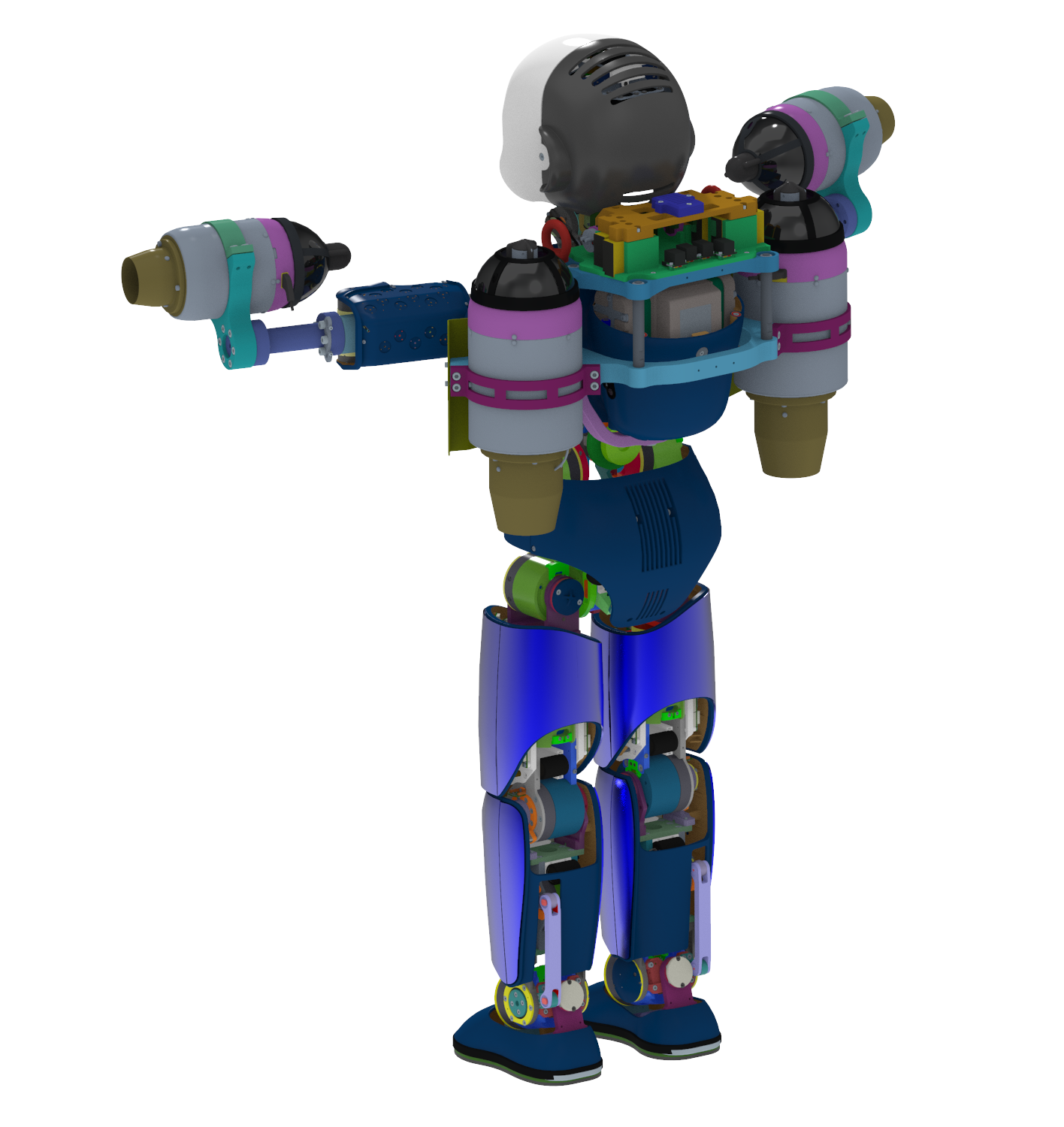}
\caption{iRonCub: the jet-powered humanoid robot. The jets are attached to the arms and the chest and exert \SI{160}{N} and \SI{220}{N} respectively.}
\label{fig:ironcub}
\end{figure}

Trajectory planning for legged robots is often addressed by using  the \textit{floating-base} formalism, which assumes that none of the robot's links has an \textit{a priori} constant position-and-orientation with respect to an inertial reference frame. 
Planning trajectories for complex legged systems using this formalism, however, is not an easy task. Most strategies are based on optimisation techniques that, dealing with issues such as the high-dimensionality of the nonlinear system and interactions with the environment, generate locally optimal motions.
Trajectory optimisation approaches using the full-robot dynamics often lead to large computational time, albeit they generate the complete set of inputs and interaction forces~\cite{posa2014direct}. Other approaches for the planning of legged systems as humanoid robots are based on linear and lower dimensional \textit{simplified models}. The linear inverted pendulum (LIP) model and the divergent component of motion (DCM) are two of these~\cite{kajita2003biped, takenaka2009real}. The resulting optimisation methods are usually faster but  not able to exploit the full robot dynamics and  kinematics constraints.  The \textit{centroidal momentum dynamics}~\cite{orin2013centroidal} is a widely used  model for robot dynamics under external forces since it projects the effects of the base and joint motions in a 6-dimensional space. The use of these reduced models improves the problem tractability while catching most of the robot governing dynamics for real applications~\cite{romualdiCentroidalMPC}. A different family of trajectory optimisation algorithms is the so-called \textit{kino-dynamic} planners. These strategies use the centroidal momentum along with the full kinematics, bounding the effects of the contact forces to the robot geometry~\cite{Dai2014, Dafarra2020Whole-BodyApproach, herzog2016structured}. \looseness=-1

Trajectory optimisation strategies for multimodal robots are still at an early stage and no general and unified approach exists to the best of the authors' knowledge. This paper takes a step in this direction by proposing a unified approach for the trajectory planning of multimodal robots expressing a degree of terrestrial and aerial locomotion. More precisely, the contribution of the paper is the design of a trajectory optimisation framework that optimizes the robot centroidal momentum dynamics under the influence of contact and thrust forces having nonlinear dynamics. We propose to incorporate the thrust dynamics along with the centroidal dynamics and the full kinematics of the robot, framing the problem as a kino-dynamic planner. The overall output of the proposed approach is a set of feasible trajectories for the robot joints, contact forces, and thrust throttles. To solve the associated optimal trajectory generation problem, we employ ADAM, a custom-made open-source library that implements a collection of algorithms for calculating rigid-body dynamics using CasADi. The overall approach is tested using iRonCub, a multimodal flying humanoid robot that can walk and fly thanks to model jet turbines installed on the robot.

The paper is organized as follows. Sec.~\ref{sec:background} introduces the notation and recalls some concepts about floating-base systems.  Sec.~\ref{sec:method} presents the  models that enable the presented trajectory optimisation framework. Sec. \ref{traj-opt} presents the trajectory optimisation formulation by specifying constraints and cost function of the associated optimal problem. Sec.\ref{sec:results} presents validation results on the iRonCub, a flying humanoid robot. Sec~\ref{sec:conclusions} closes the paper with the conclusion perspectives.\looseness=-1

%% file: tex/background.tex
\section{Background}
\label{sec:background}

\subsection{Notation}
\begin{itemize}
    \item $\mathbf{I}_n$ and $\mathbf{0}_n$ denote the $n \times n$ identity and zero matrix;
    \item $\asFrame{I}$ denotes the inertial frame;
    \item ${}^\asFrame{A}p_\asFrame{C}$ represents the vector connecting the origin of the frame $\asFrame{A}$ to the origin of the frame $\asFrame{C}$, expressed in $\asFrame{A}$;
    \item ${}^\asFrame{A}R_\asFrame{B} \in SO(3)$ and ${}^\asFrame{A}H_\asFrame{B} \in SE(3)$ are the rotation matrix and the homogeneous transform matrix;
    \item $\hat{\cdot}$ defines the \textit{hat operator}: $\R^3 \rightarrow \mathfrak{so}(3)$, where $\mathfrak{so}(3)$ is the set of skew-symmetric matrices. $\hat{x}y = x \times y$, where $\times$ is the cross product operator in $\R^3$. $x \times = S(x)$
    \item ${}^\asFrame{A}f_\asFrame{B} \in \R^3$ and ${}^\asFrame{A}\tau_\asFrame{B} \in \R^3$ are the force and torque acting on the rigid body on the frame $\asFrame{B}$ and expressed in $\asFrame{A}$;
    \item $\mathrm{f}_\asFrame{B} = \begin{bmatrix}
    {}^\asFrame{A}f_\asFrame{B}^\top & {}^\asFrame{A}\tau_\asFrame{B}^\top \end{bmatrix}^\top$ identifies the wrench acting on a point of the rigid body; 
    \item $g$ is the gravity vector expressed in the inertial frame $\asFrame{I}$.
\end{itemize}

\subsection{Floating-base modeling}

A multimodal robot can be modeled as a multi-body system composed of $n+1$ rigid bodies -- called links -- connected by $n$ joints with one degree of freedom. Using of the \textit{floating base} formalism, we define the \textit{configuration space} as $\mathbb{Q} \in \R^3 \times SO(3) \times \R^n$.  More precisely, 
$q=({}^\asFrame{I}p_\asFrame{B}, {}^\asFrame{I}R_\asFrame{B}, s) $ defines a specific robot configuration.
The velocity of the system belongs to the set $\mathbb{V} \in \R^3 \times \R^3 \times \R^n$. An element of $\mathbb{V}$ is given by $\nu = ({}^\asFrame{I}\dot{p}_\asFrame{B}, {}^\asFrame{I}\omega_\asFrame{B}, \dot{s})$, where $({}^\asFrame{I}\dot{p}_\asFrame{B}, {}^\asFrame{I}\omega_\asFrame{B})$ are the linear and angular velocity of the base frame $\asFrame{B}$ w.r.t. the inertial $\asFrame{I}$ and $\dot{s}$ are the joint velocities. The relationship $\dot{{}^\asFrame{I}R_\asFrame{B}} = S({}^\asFrame{I}\omega_\asFrame{B}) {}^\asFrame{I}R_\asFrame{B} $ is then satisfied.
By applying the Euler-Poincaré formalism, the equation of motion of a system exchanging $n_b$ wrenches with the environment results in
\begin{equation}
    M(q) \dot{\nu} + C(q, \nu) \nu + G(q) = \begin{bmatrix} 0_{6 \times 1} \\ \tau \end{bmatrix} + \sum _{i=1} ^{n_b} J^\top _i \mathrm{f}_i,
    \label{system_dyn}
\end{equation}

\noindent 
where $M, C \in \R ^{(n+6) \times (n+6)}$ are the mass and Coriolis matrix (accounts for the Coriolis and centrifugal terms), $G \in \R ^{n+6}$ is the gravity vector, $\tau \in \R ^{n}$ are the internal actuation torques, $\mathrm{f}_i$ is the $i$th of the $n_b$ external wrench applied on the robot. The application point of the wrench $\mathrm{f}_i$ is assumed to be the origin of a frame $\asFrame{C}_i$ attached to the link where the wrench acts. The wrench $\mathrm{f}_i$ is expressed in the inertial frame $\asFrame{I}$. The jacobian $J_i(q)$ maps the the system velocity $\nu$ to the velocity $({}^\asFrame{I}\dot{p}_{\asFrame{C}_i}, {}^\asFrame{I}\omega_{\asFrame{C}_i})$ of the frame $\asFrame{C}_i$.

%% file: tex/method.tex
\section{Models for Trajectory Optimisation}
\label{sec:method}

This section introduces the  multimodal robot models that are leveraged in the predictive trajectory layer of Sec. \ref{traj-opt}. 

\subsection{Centroidal Momentum Dynamics}
\label{subsec:centroidal}

Eq. \eqref{system_dyn} is the rigid multi-body model of the underlying robot and characterises the time-evolution of both the robot joints $s$ and the floating base $({}^\asFrame{I}p_\asFrame{B}, {}^\asFrame{I}R_\asFrame{B})$. Planning robot trajectories using  \eqref{system_dyn}, however, leads to \emph{large} optimisation problems often having very high computational time, which also depend on sets of gains not easy to tune. For this reason, robot trajectory optimisation often leverages system representations different from \eqref{system_dyn} with the aim of dealing with models leading to simpler and faster optimisation problems to solve. A possible approach is to use \emph{reduced} robot models that follow from \eqref{system_dyn} but capture the time evolution of a reduced set of the robot state variables. One of these reduced models is the so-called \textit{centroidal momentum}, namely, a six-dimensional quantity describing the time evolution of the total robot momentum expressed at a proper frame of reference~\cite{orin2013centroidal}. \looseness=-1

Let $\asFrame{G}$ be a frame having the origin at the robot CoM  $x \in \mathbb{R}^3$ and the orientation of the inertial frame $\asFrame{I}$. The \textit{centroidal momentum} is then denoted by
\begin{equation}
    {}^\asFrame{G}h = \begin{bmatrix} {}^\asFrame{G}h_p \\ {}^\asFrame{G}h_\omega \end{bmatrix} \in \R^6,
\end{equation}
and is given by the compound linear and angular momentum of each link expressed in the frame $\asFrame{G}$. 

The centroidal momentum dynamics is the sum 
of the external forces and moments that act on the system
\begin{equation}
    {}^\asFrame{G}\dot{h} = \sum_{i} ^{n_b} 
    \begin{bmatrix} \mathbf{I}_3 & \mathbf{0}_3 \\ 
    S(p_{i} - x) & \mathbf{I}_3\end{bmatrix} \mathrm{f}_i + m \bar{g},
    \label{eq:centroidal_dyn}
\end{equation}
being $\bar{g} = \begin{bmatrix} g^\top & 0_{1 \times 3} \end{bmatrix}^\top$, $m$ the total mass of the robot, and $p_i$ the origin of the $i$-th of one of the $n_b$ bodies on which the wrenches are applied.

Moreover, the following relationship holds
\begin{equation}
    \label{eq:centroidal}
    {}^\asFrame{G}h = J_\text{CMM}(q) \nu,
\end{equation}
where $J_\text{CMM}$ is the \textit{Centroidal Momentum Matrix} (CMM), mapping the robot velocity to the centroidal momentum~\cite{orin2013centroidal}.
We recall that the linear momentum relates with the robot CoM velocity depending on the total mass $m$ as ${}^\asFrame{G}h_p = m \dot{x}$. %

Under the assumption that the joint quantities can be chosen at will, Eq.~\eqref{eq:centroidal_dyn} expresses the entire robot dynamics, which makes Eq.~\eqref{system_dyn} not necessary for trajectory optimisation purposes. For the sake of clarity the superscripts $\asFrame{G}$, $\asFrame{I}$, and $\asFrame{B}$ will be omitted in the following sections.

\subsection{Models for the actuation of aerial locomotion}

Multimodal robots express both aerial and terrestrial locomotion. Hence, we assume that the robot is powered by $n_\text{jets}$ thrust forces for achieving aerial locomotion -- see Fig~\ref{fig:ironcub} for an example. Every thrust force has an intensity denoted by $T_i \in \R$ and its direction and application point are defined by proper frames associated to each thrust vector.  In particular, we assume that the $i$-th thrust generates a pure force, applied at the point $p_i$, given by
\begin{equation}
f_i = R_i(q) \begin{bmatrix}
0 \\ 0 \\ -T_i
\end{bmatrix}, \forall i \in [0, n_\text{jets}],   
\end{equation}
where $R_i(q)$ represents the orientation of the aforementioned frame.

Assuming that thrust forces apply pure forces leads to the following \textit{centroidal momentum} dynamics expression
\begin{equation}
    \dot{h} = \sum_{i} ^{n_\text{jets}} 
    \begin{bmatrix} \mathbf{I}_3 \\ 
    S(p_{i} - x) \end{bmatrix} f_i + m \bar{g}.
\label{eq:centroidal_dyn_jet}
\end{equation}

In many cases, thrust forces cannot be chosen at will~\cite{Pucci2017MomentumRobot, Nava2018PositionRobot,LErario2020ModelingRobotics}, and additional models for characterising thrust dynamics shall be taken into account.
By neglecting the subscript $i$, we assume that the thrust dynamics is described by a second-order nonlinear model
\begin{equation}
    \ddot{T} = f(T, \dot{T}) + g(T, \dot{T}, u).
\end{equation}

For example, jet turbines can generate thrust forces for aerial humanoid robots. Using data-driven approaches, one can show that jet dynamics are given by \cite{LErario2020ModelingRobotics}
\begin{subequations}
    \begin{align}
        \begin{split}
        f(T, \dot{T}) & = K_T T {+} K_{TT} T^2 {+} K_D \dot{T} \\ & +  K_{DD}\dot{T}^2 {+} K_{TD} T \dot{T} + c, 
        \end{split} \\
    	 g(T, \dot{T}) & = (B_U + B_T T + B_D \dot{T}) (u + B_{UU}u^2), 
     \end{align}
 \label{eq:jet_dynamics_fun}
\end{subequations}
where $K$ and $B$ are a set of parameters that relate the evolution of the jet state $[T, \dot{T}]^\top$ to the jet throttle $u$.

\subsection{Models for the actuation of terrestrial locomotion}
\label{subsec:act-terrestrial}
We assume that the robot can achieve terrestrial locomotion by making and breaking contact with the environment. 
So, we assume that the robot may be in contact with a rigid environment at $n_c$ distinct locations.
Furthermore, we assume that the robot makes \emph{surface} contacts, and the contact surface can be approximated with a rectangle. Consequently, the net wrench resulting from the interaction between the robot and the environment at the contact location can be described by four pure forces acting at the corner of the rectangle. %
In light of the above, the \textit{centroidal dynamics} under the presence of contact forces and zero thrust writes
\begin{equation}
    \dot{h} = \sum_j ^{n_c}
    \begin{bmatrix} \mathbf{I}_3  \\ 
    S(p_j - x) \end{bmatrix} f_j + m \bar{g},
    \label{eq:centroidal_dyn_contacts}
\end{equation}
where $p_j$ is the position of the vertex $j$ and $f_j$ is the pure force acting on it.

The contact force is assumed to be non-null when the point is in contact with the surface. This condition is described by the so-called \textit{complementary condition}
\begin{equation}
    \mathbf{d}(p_j) \mathbf{n}(p_j) ^\top f_j = 0,
    \label{eq:complementarity_condition}
\end{equation}
where $\mathbf{d}(p_j)$ returns the distance of the point $p_j$ from the surface and $\mathbf{n}(p_j)$ computes the normal of the surface at the point. 
When the robot makes a \emph{stable} contact with the environment, the contact does not slide, i.e. when the contact force is non-zero, the tangential motion of the contact point is null. This condition is called \textit{planar complementarity condition} and is given by
\begin{equation}
    (\mathbf{t}(p_j)^\top\dot{p}_j) (\mathbf{n}(p_j) ^\top f_j) = 0,
      \label{eq:planar_complementarity_condition}
\end{equation}
where $\mathbf{t}$ returns the tangential component of the surface at $p_j$.\looseness=-1

The contact force belongs to a second-order cone called \textit{friction cone}~\cite{Lobo1998ApplicationsProgramming}. Nevertheless, the friction cone is oftentimes approximated with a set of linear inequalities of the type
\begin{equation}
    - \mu \ \textbf{n} (p_j) ^\top f_j \le \textbf{t}(p_j) ^\top f_j \le \mu \ \textbf{n}(p_j) ^\top f_j,
\end{equation}
where  $\textbf{t}(p_j) ^\top f_j$ represents the tangential component of the contact forces to the surface. This constraint is re-writable in matrix form as $A f_j \le b$. 

Summing up the contributions of the contact forces~\eqref{eq:centroidal_dyn_contacts} and of the thrust  forces~\eqref{eq:centroidal_dyn_jet}, \eqref{eq:centroidal_dyn} is rewritten as
\begin{equation}
\begin{split}
        \dot{h} &= \sum_{i} ^{n_\text{jets}} 
    \begin{bmatrix} \mathbf{I}_3 \\ 
    S(p_{i} - x) \end{bmatrix} R_i \begin{bmatrix}
0 \\ 0 \\ -T_i \end{bmatrix} \\ &+ \sum_j ^{n_v}
    \begin{bmatrix} \mathbf{I}_3  \\ 
    S(p_j - x) \end{bmatrix} f_j + m \bar{g}.
\end{split}
\label{eq:centroidal_dyn_total}
\end{equation}

\section{Trajectory optimisation}
\label{traj-opt}

This section presents the trajectory optimisation algorithms for multimodal robots characterised by the models presented above. 
The main aim is to cast an optimal control problem whose solution represents feasible robot trajectories.

The optimisation algorithms presented below make use of the centroidal momentum described in section~\ref{subsec:centroidal}. 
The centroidal momentum dynamics relates the external wrenches acting on the robot to the robot's center of mass (i.e. linear momentum) and \emph{average angular motions} (i.e. angular momentum). Being the centroidal momentum related to the robot kinematics by~\eqref{eq:centroidal}, momentum-based trajectory planning allows us to enforce joint constraints, 
end-effector targets, and kinematic penalties. In fact, we can take into account the full robot kinematics relating the robot CoM, contact and thrust forces, and the centroidal momentum to the robot configuration $q$ and its velocity $\nu$.

The optimisation problem for momentum-based trajectory planning is then given by the components  presented next:
\begin{itemize}
    \item a cost function $J$ that accounts for regularization tasks and desired final state;
    \item a set of constraints enforcing the robot dynamics, the thrusts, the robot kinematics, and the contacts;
    \item a set of constraints on initial and final centroidal momentum quantities. These play a pivotal role: they are in charge of defining initial and final equilibrium, in addition to initial and \textit{desired} final CoM position.
\end{itemize}

The problem is transcribed using a direct \textit{multiple-shooting} method. The trajectories of the optimisation variables are discretized with $N$ knots, with a time interval $\Delta t$. %

The base rotation $R$ is vectorized using the quaternion representation. Hence, from now on, $q := [p^\top \ \rho^\top \ s^\top]^\top$, where $\rho \in \mathbb{H}$ is a unit quaternion.

The decision variables of the nonlinear problem are:
\begin{itemize}
    \item CoM position, velocity and acceleration $x$, $\dot{x}$, $\ddot{x}$;
    \item the angular momentum and its time derivative $h_\omega$, $\dot{h}_\omega$;
    \item the robot configuration and its velocity $q$, $\nu$, $\dot{s}$;
    \item the contact forces and points $f$, $\dot{f}$, $p$;
    \item the thrust quantities and throttle $T$, $\dot{T}$, $\ddot{T}$, $u$;
    \item the total time $N \Delta t $.
\end{itemize}

\subsection{Constraints}

Constraints ensure that the optimised solution meets the defined physical description, e.g. robot dynamics consistency, contact forces in the friction cones and fulfilling the complementarity conditions, etc, at each knot $k$.

\subsubsection{Centroidal dynamics}

The six-dimensional centroidal dynamics taking into account thrust and contact forces -- see \eqref{eq:centroidal_dyn_total} -- is enforced by the constraint

\begin{equation}
\begin{split}
    \begin{bmatrix}
    \ddot{x} [k] m \\ \dot{h}_\omega[k]
    \end{bmatrix} = \sum_i ^{n_c} \begin{bmatrix} \mathbf{I}_3 \\ 
    S(p_i[k] - x[k]) \end{bmatrix} f_i[k] \\ +
    \sum_{j} ^{n_\text{jets}} 
    \begin{bmatrix} \mathbf{I}_3 \\ 
    S(p_j[k] - x[k]) \end{bmatrix}R_j[k] \begin{bmatrix}
0 \\ 0 \\ -T_i[k] \end{bmatrix}.
\end{split}
\label{eq:mom_constr}
\end{equation}

\subsubsection{Centroidal Momentum and Kinematics}

The time evolution of the centroidal momentum  is directly related to the robot kinematics -- see \eqref{eq:centroidal}. For instance, the linear robot momentum is given by ${}^\asFrame{G}h_p = m \dot{x}$ and, in turn, the robot CoM $x$ is a function of the robot  configuration, namely
\begin{equation}
    x[k] = \text{CoM}(q[k]).
    \label{eq:com_constr}
\end{equation}

Likewise, the angular momentum also depends upon the robot configuration and velocity. By isolating the last three equations of \eqref{eq:centroidal}, one then obtains that the angular centroidal momentum is constrained by
\begin{equation}
    h_\omega[k] = \begin{bmatrix}\mathbf{0}_{3} & \mathbf{I}_{3}\end{bmatrix} J_\text{CMM}(q[k]) \nu [k].
    \label{eq:cmm_constr}
\end{equation}

\subsubsection{Contact points and forces}
Robot contact points  must satisfy the robot kinematics. The following equality then ensures that the contact position $p_i$ equals the point computed using the robot forward kinematics of the $i$-th contact frame %
\begin{equation}
    p_i[k] = \text{fk}_i(q[k]).
    \label{eq:point_fk}
\end{equation}

Assuming rigid contacts presented in Sec. \ref{subsec:act-terrestrial}, one has that the contact should not penetrate the surface and that normal component of the reaction force to the surface shall be non-negative, i.e. we obtain the following inequality constraint
\begin{equation}
    \mathbf{d}(p_i[k]) \ge 0, \quad \mathbf{n}(p_i[k]) ^\top f_i[k] \ge 0.
    \label{eq:contact_non_negativity}
\end{equation}

Every force should lie inside the linearized friction cone
\begin{equation}
    A f_i[k] \le b,
    \label{eq:friction_cone}
\end{equation}
where $A$ and $b$ depend on the static friction coefficient $\mu$.

To determine the contact conditions, one may use the complementarity conditions~\eqref{eq:complementarity_condition} and~\eqref{eq:planar_complementarity_condition}. When used in an optimisation algorithm as a constraint, however,~\eqref{eq:complementarity_condition}, \eqref{eq:planar_complementarity_condition} may lead to numerical issues: the constraint Jacobian may get singular, violating the linear independence constraint qualification (LIQC) that many nonlinear solver algorithms exploit to find optimal solutions~\cite{Betts_practical}. For this reason, the complementarity condition~\eqref{eq:complementarity_condition} is implemented in its relaxed form, which leads to fewer numerical problems
\begin{equation}
    \mathbf{d}(p_i[k]) \mathbf{n}(p_i[k]) ^\top f_i[k] \le \epsilon_n[k],
    \label{eq:relaxed_complementarity}
\end{equation}
where $\epsilon_n$ is a so-called bounded \textit{panic} variable that is supposed to be as small as possible. Hence, we include $\epsilon_n$ in the cost function so as its 2-norm is minimised. 

The planar complementarity condition is implemented in its discretized and \textit{relaxed} form, which writes
\begin{equation}
    -\epsilon_t[k] \le \mathbf{t}(p_i[k])\Big(p_i[k+1] - p_i[k]\Big) \mathbf{n}(p_i[k]) ^\top f_i[k] \le \epsilon_t[k], 
    \label{eq:planar_relaxed_complementarity}
\end{equation}
where, in the same spirit of~\eqref{eq:relaxed_complementarity}, we make use of a bounded panic variable $\epsilon_t$ that is added in the cost function.

\subsubsection{Jet location and dynamics}
Thrust forces are associated with frames, so we add the following constraint
\begin{equation}
    (p_j, R_j) = \text{fk}_j (q[k]).
\label{eq:jet_fk}
\end{equation}

The thrust dynamics, instead, is enforced with
\begin{equation}
    \ddot{T}[k] = f(T[k], \dot{T}[k]) + g(T[k], \dot{T}[k], u[k]),
    \label{eq:jet_dynamics}
\end{equation}
where $f(T, \dot{T})$ and $g(T, \dot{T}, u)$ are the functions~\eqref{eq:jet_dynamics_fun} regulating the thrust dynamics. 

\subsubsection{Variables integration} The evolving optimisation variables are projected along the horizon using backward Euler method, which improves numerical stability~\cite{Betts_practical}. 
The momentum dynamics evolves as
\begin{subequations}
\begin{align}
    h_\omega[k+1] &= h_\omega[k] + \dot{h}_\omega[k+1] \Delta t, \\
    x[k+1] &= x[k] + \dot{x}[k+1] \Delta t, \\
    \dot{x}[k+1] &= \dot{x}[k] + \ddot{x}[k+1] \Delta t.
\end{align}
\label{eq:momentum_integration}
\end{subequations}

The robot configuration, in the same fashion, evolves as
\begin{subequations}
\begin{align}
    p[k+1] &=p[k] + \dot{p}[k+1] \Delta t,\\
    \rho[k+1] &= \omega[k + 1] \Delta t \oplus \rho[k] , \label{eq:rotation_integration_with_exp_map} \\
    s[k+1] &= s[k] + \dot{s}[k+1] \Delta t,
\end{align}
\label{eq:configuration_integration}
\end{subequations}
where the operator $\oplus$ in~\eqref{eq:rotation_integration_with_exp_map} redefines the addition operator for the rotation group. In particular, $\oplus$ implements the exponential operator $\omega \Delta t \oplus \rho = \text{Exp}(\omega \Delta t) \rho$, ensuring the membership in the SO(3) group of the variable representing the rotation~\cite{sola2018micro}.

Finally, the thrust dynamics evolves as
\begin{subequations}
\begin{align}
    T[k+1] = T[k] + \dot{T}[k+1] \Delta t, \\
    \dot{T}[k+1] = \dot{T}[k] + \ddot{T}[k+1] \Delta t.
\end{align}
    \label{eq:jet_integration}
\end{subequations}
We introduce the variation of the contact forces as an optimisation variable, which integrates as
\begin{equation}
    f[k+1] = f[k] + \dot{f}[k+1] \Delta t,
\label{eq:contact_forces_integration}
\end{equation}
and allows us to obtain smoother contact forces profiles.

\subsubsection{Box constraints}
We bound the optimisation variables to their physical limitations using box constraints. In this sense, joints values are bounded inside their range of motion and have a finite variation
\begin{subequations}
\begin{align}
    s_\text{min} \leq &s[k] \leq s_\text{max}, \\
    \dot{s}_\text{min} \leq &\dot{s}[k] \leq \dot{s}_\text{max}.
\end{align}
\label{eq:joint_box_constr}
\end{subequations}
Based on the jet specifications, we limit also the jet throttle
\begin{equation}
    u_\text{min} \leq u[k] \leq u_\text{max}.
    \label{eq:jet_box_constr}
\end{equation}

\subsubsection{Boundary constraints}
Initial and final constraints play a key role. We specify the task, mostly defined by the desired final CoM position, through the following constraints:
\begin{itemize}
    \item Initial constraints: they force the system at equilibrium and the initial CoM position variable at $x_0$
    \begin{subequations}
    \begin{align}
        x[0] &= x_0, \\
        h_\omega[0] &= 0, \dot{x}[0] = 0.
    \end{align}
    \end{subequations}
    \item Final constraints: they force the final CoM position variable at a desired $x_N$
    \begin{align}
        x[N] &= x_N.
    \end{align}
\end{itemize}

\subsection{Cost function}

Whilst the final constraints define the final goal, the objective function shapes \textit{how} this final goal is reached. 
So, the cost function to be minimised is defined as 
\begin{equation}
    J = l_N + \sum _{k=0} ^{N-1} l[k] \Delta t.
\end{equation}

At each knot $k$, the running cost $l[k]$ accounts for regularization tasks 
and it is defined as the weighted 2-norm of the optimisation variables, namely
\begin{align*}
    l[k]  & =  \| \dot{x}[k] \|_{w_x} + \| \ddot{x}[k] \|_{w_{\dot{x}}}  + \| h_{\omega}[k] \|_{w_h} + \| \dot{h}_{\omega}[k] \|_{w_{\dot{h}}} \\
    & + \| \dot{s}[k] \|_{w_s} {+} \| \dot{v}[k] \|_{w_v} {+} \| s[k] - \bar{s}\|_{w_{\bar{s}}} {+} \| \epsilon_n[k] {+} \epsilon_t[k] \|_{w_{\epsilon}} \\
    & + \| \dot{f}[k] \|_{w_{\dot{f}}} + \| f[k] \|_{w_f}  + \| \dot{T}[k] \|_{w_{\dot{T}}}  + \| u[k] \|_{w_u}, 
\end{align*}
while the final cost $l_N$ drives the momemtum dynamics to the equilibrium
\begin{equation*}
    l_N  =  \| \dot{x}[N] \|_{\bar{w}_x} + \| \ddot{x}[N] \|_{\bar{w}_{\dot{x}}}  + \| h_{\omega}[N] \|_{\bar{w}_h} + \| \dot{h}_{\omega}[N] \|_{\bar{w}_{\dot{h}}}
\end{equation*}
where $w$, $\bar{w}$ are weights for each term in the cost function and $\bar{s}$ the desired joint postural configuration.

\subsection{Complete Optimisation Problem}
Summing up, the optimisation is
\begin{equation}
\begin{aligned}
\min_{\chi} \quad & l_N + \sum _{k=0} ^{N-1} l[k] \Delta t \\
\textrm{s.t.} \quad & \text{Centroidal dynamics~\cref{eq:mom_constr}}, \\
                    & \text{Centroidal momentum~\cref{eq:com_constr,eq:cmm_constr}},    \\
                    & \text{Contact points and forces~\cref{eq:point_fk,eq:contact_non_negativity,eq:friction_cone,eq:relaxed_complementarity,eq:planar_relaxed_complementarity}}, \\
                    & \text{Jet location and dynamics~\cref{eq:jet_fk,eq:jet_dynamics}}, \\
                    & \text{Box constraints~\cref{eq:joint_box_constr,eq:jet_box_constr}}, \\
                    & \text{Variables integration~\cref{eq:momentum_integration,eq:configuration_integration,eq:jet_integration,eq:contact_forces_integration}}, \\
                    & \chi[0] = \chi_0, \\
                    & \chi[N] = \chi_N.
\end{aligned}
\label{eq:complete_opti_problem}
\end{equation}
where $\chi$ collects all the optimisation variables.

%% file: tex/results.tex
\section{Validation on an Aerial Humanoid Robot}
\label{sec:results}

\begin{figure*}[tpb]
\vspace{0.2cm}
    \centering
    \begin{myframe}{Take-off}
    \subfloat[Snapshots of the robot take-off.] {%
    \begin{minipage}{\linewidth} \label{fig:take_off_snapshots}
     \includegraphics[width=0.33\textwidth]{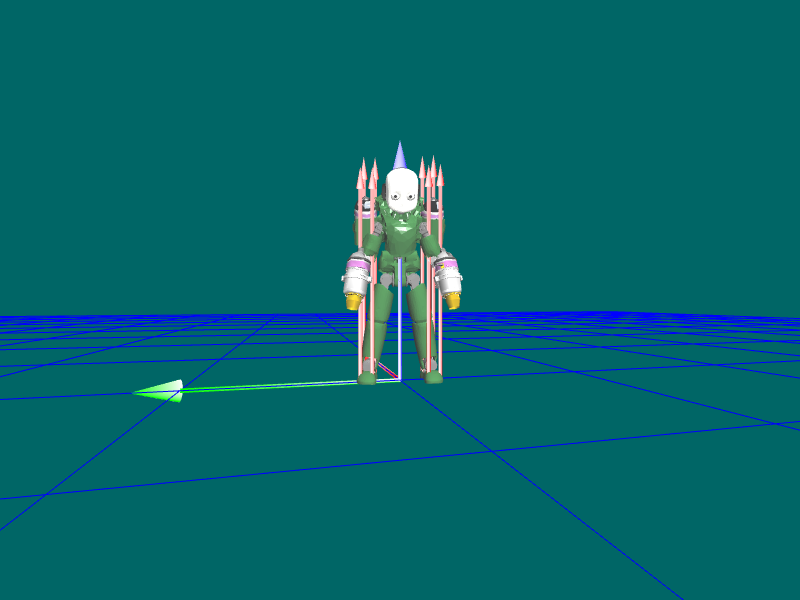}
     \includegraphics[width=0.33\textwidth]{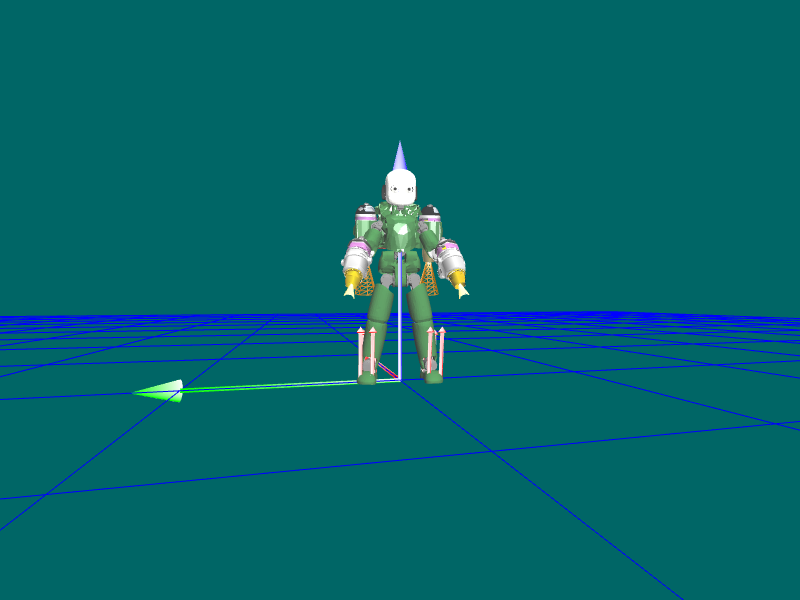}
     \includegraphics[width=0.33\textwidth]{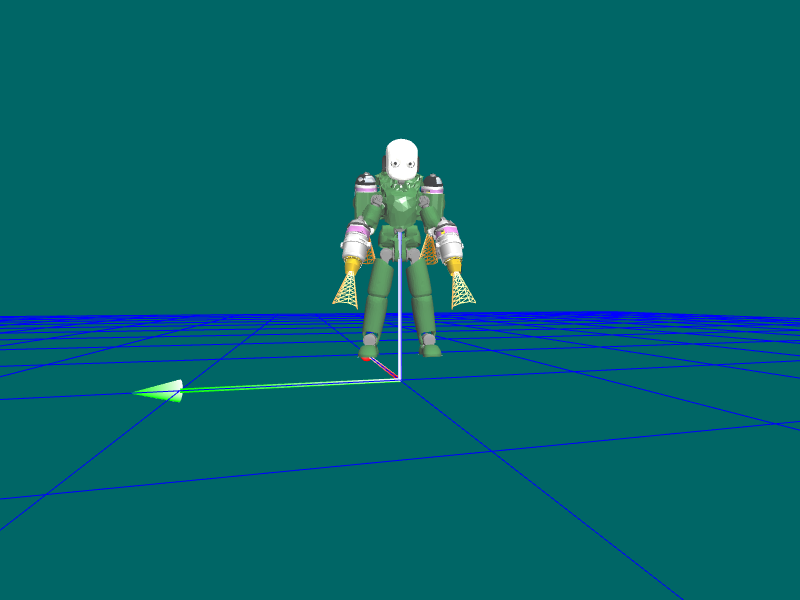}
     \end{minipage}} \par
     \subfloat[Trajectories of throttle, thrust and total vertical component of the contact forces during the take-off. The red band represents the flight phase.]{%
     \begin{minipage}{\linewidth} \label{fig:take_off_plots}
     \includegraphics[width=0.33\textwidth]{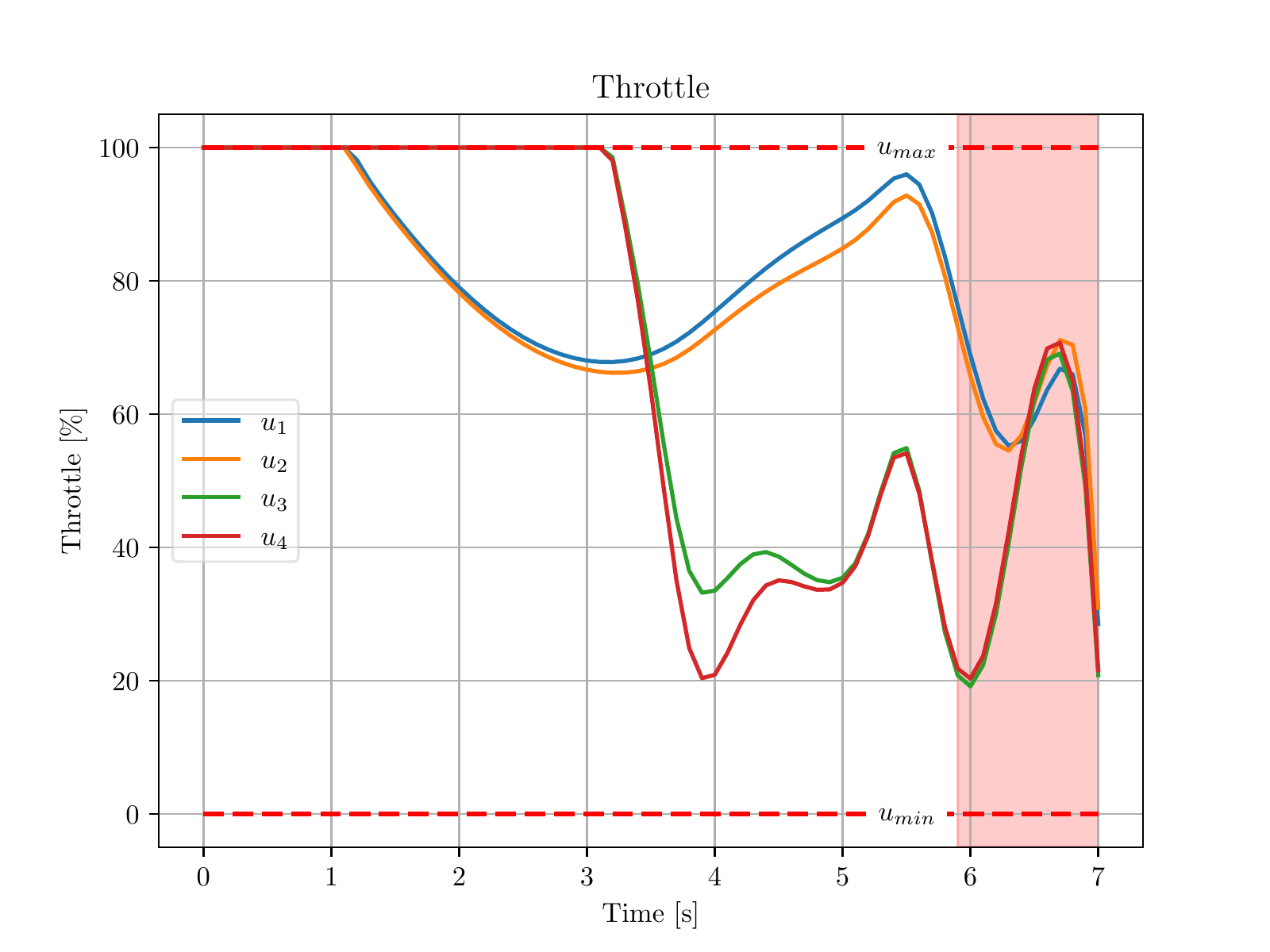}
     \includegraphics[width=0.33\textwidth]{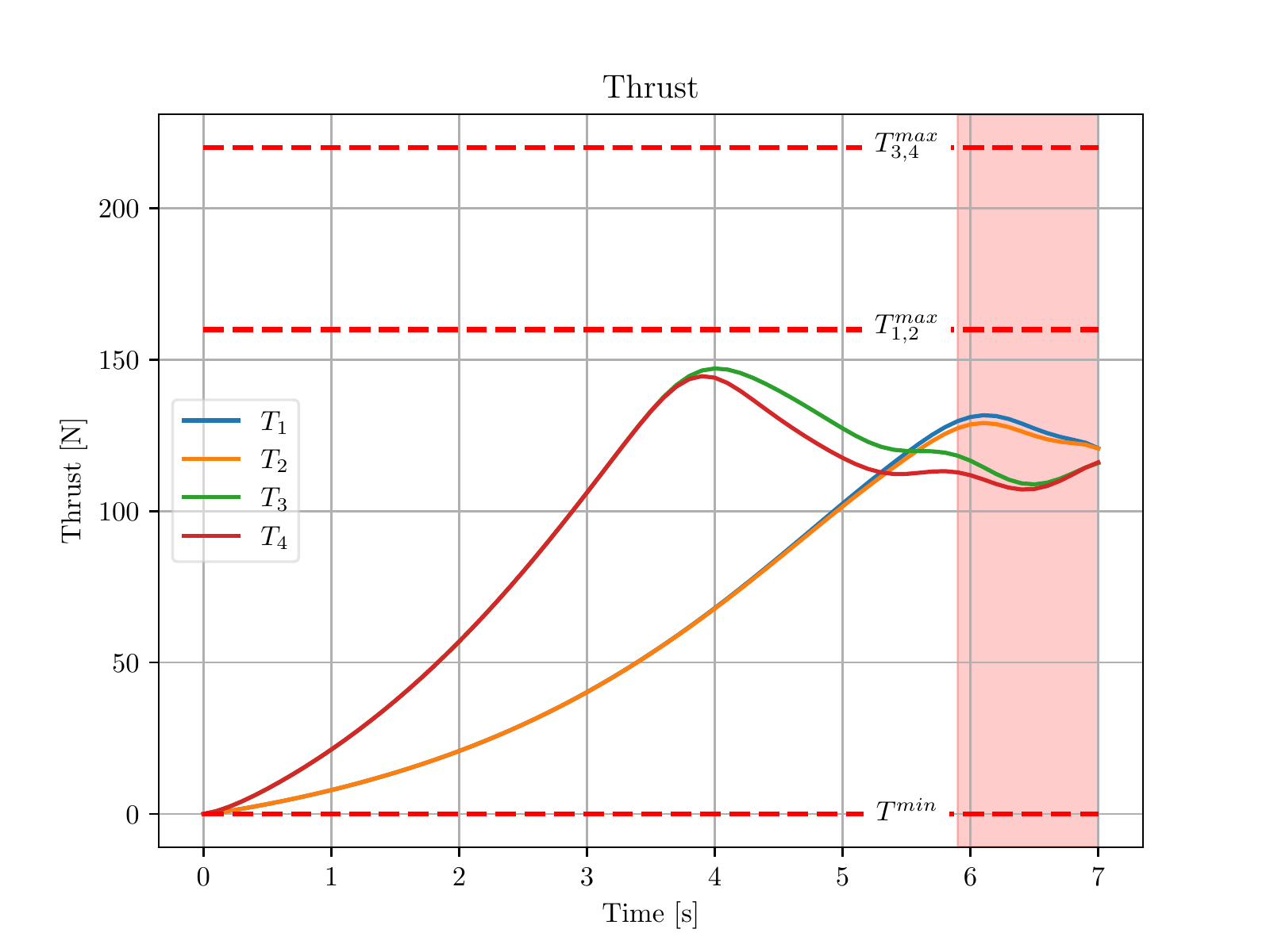}
     \includegraphics[width=0.33\textwidth]{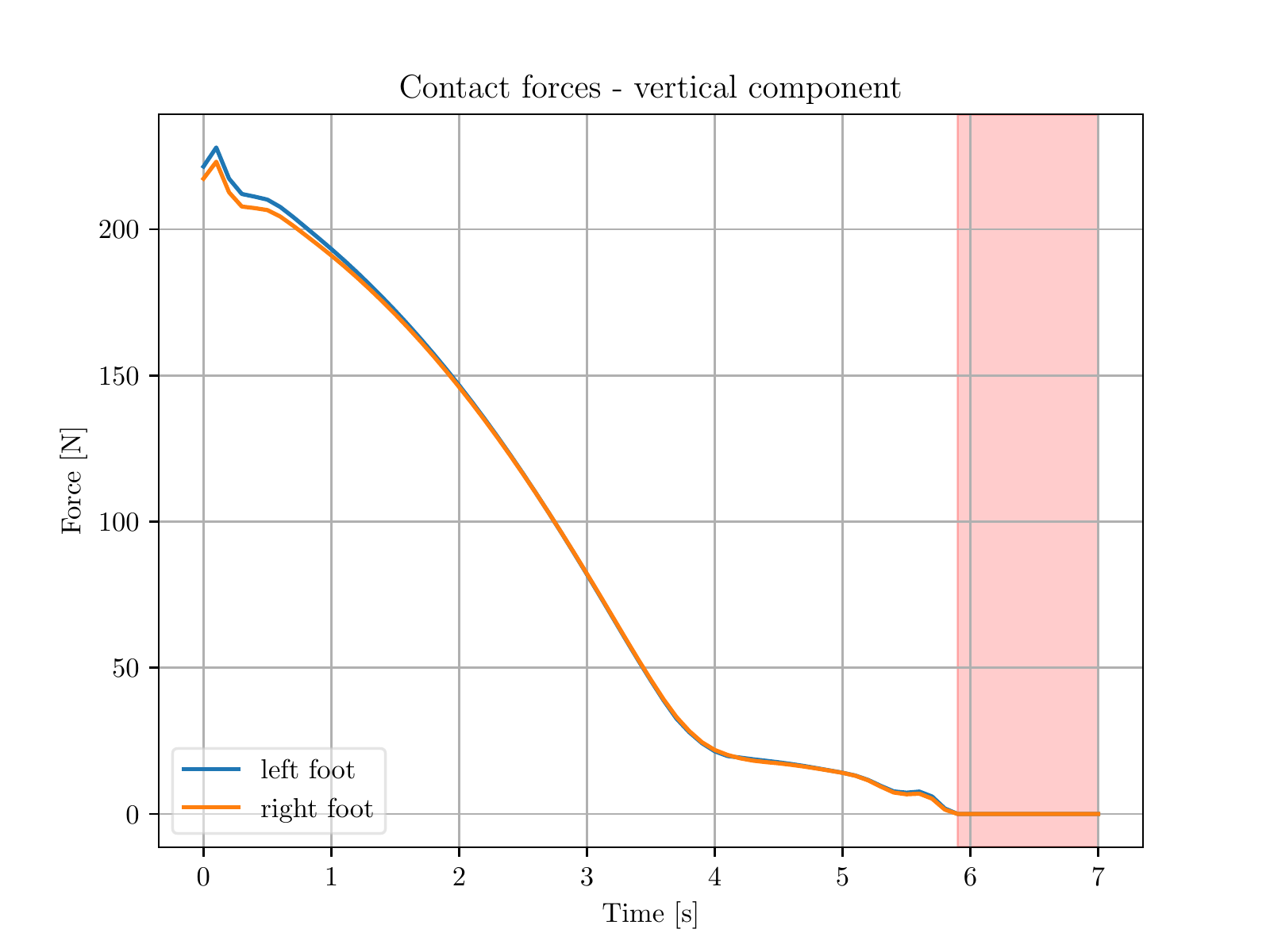}
     \end{minipage}}
    \end{myframe}
    \caption{Snapshots and trajectories of the robot during the take-off.} 
\end{figure*}

This section presents validations for the modelling and optimisation framework above using iRonCub, a flying humanoid robot that expresses terrestrial and aerial locomotion.

An approach to solving the nonlinear problem \eqref{eq:complete_opti_problem} makes use of a software framework that computes the rigid-body dynamics of the robot in a symbolic fashion, allowing its propagation in the context of direct transcription methods. For this purpose, we implemented a custom library called ADAM\footnote{\url{https://github.com/ami-iit/ADAM}} that implements a collection of algorithms for calculating rigid-body dynamics using CasADi~\cite{Andersson2019}. The optimisation problem is then transcribed using the Opti Stack and solved using IPOPT~\cite{wachter2006implementation} 3.13.4, with the MA97 linear solver.

We carry out the validation of the methodology using the dynamical model of the jet-powered version of iCub~\cite{Natale2017ICub:Child}, namely, the iRonCub. The robot possesses 23 degrees of freedom and weights \si{44}{kg}. It is equipped with 4 model jet engines, 2 on the arms and 2 on the chest, capable of exerting maximum thrust forces of $\SI{160}{N}$ and $\SI{220}{N}$ respectively. These limits are included in the problem \eqref{eq:complete_opti_problem}. The parameters of the jet system dynamics \eqref{eq:jet_dynamics_fun} are those identified in~\cite{LErario2020ModelingRobotics}.  

We validate the trajectory optimisation framework in different scenarios. First, we test the planner for vertical take-off and landing, walking to flying transition, and jumping. In all the tests, the contact sequence emerges directly from the complementarity formulation and the robot chooses the timing and the contact location freely. The set of weights used in the cost function shapes \textit{how} this task is performed and strongly depends on the task itself. The computational time that the solver spends to generate a trajectory strongly depends on the provided initial guess and goes from hundreds of seconds up to several minutes. Software implementation and results can be found at~\url{https://github.com/ami-iit/paper_lerario_2022_planning-multimodal-locomotion}.

\subsection{Take-off}
The first scenario consists of robot take-off. The robot starts at equilibrium and the jet thrust is set to zero. 
We set the initial CoM position at $x_0 = \SI{0.57}{m}$ and the final one at $x_0 = \SI{0.7}{m}$.  
The net-force transitions smoothly from contact forces to jet thrusts, as Fig~\ref{fig:take_off_plots} shows. The boundaries on thrust and throttle are satisfied and the jet behaviour complies with the jet dynamics. The horizon is 70 knots long. The generated trajectories are shown in Fig~\ref{fig:take_off_snapshots}. We used high weights on joint velocity to obtain a slower motion of the robot.\looseness=-1

\subsection{Vertical Take-off and Landing}
\begin{figure*}[tbp]
    \centering
    \begin{myframe}{Take-off and landing}
    \subfloat[Snapshots of the robot taking-off and landing.] {%
    \begin{minipage}{\linewidth} \label{fig:tol_snapshots}
     \includegraphics[width=0.33\textwidth]{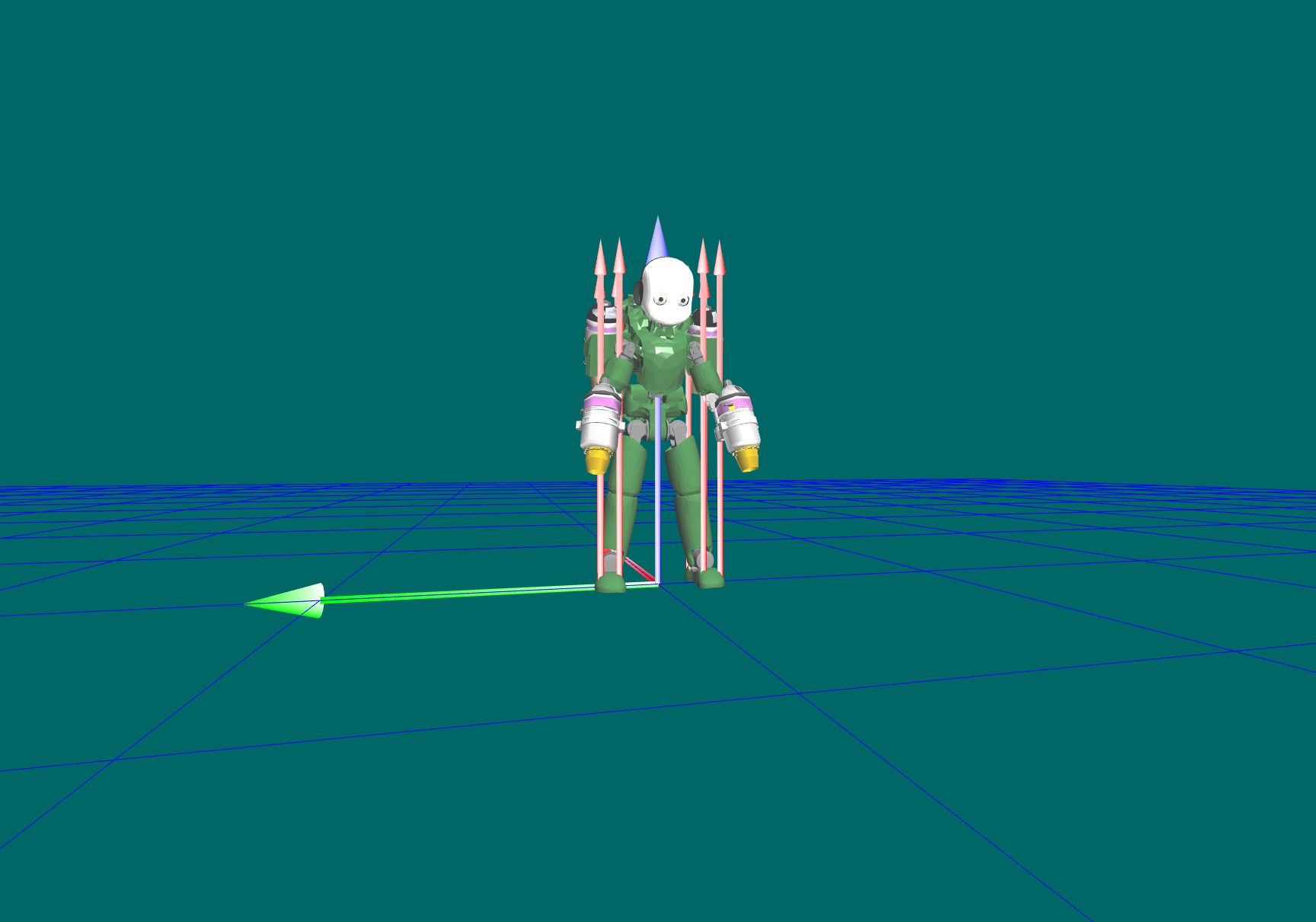}
     \includegraphics[width=0.33\textwidth]{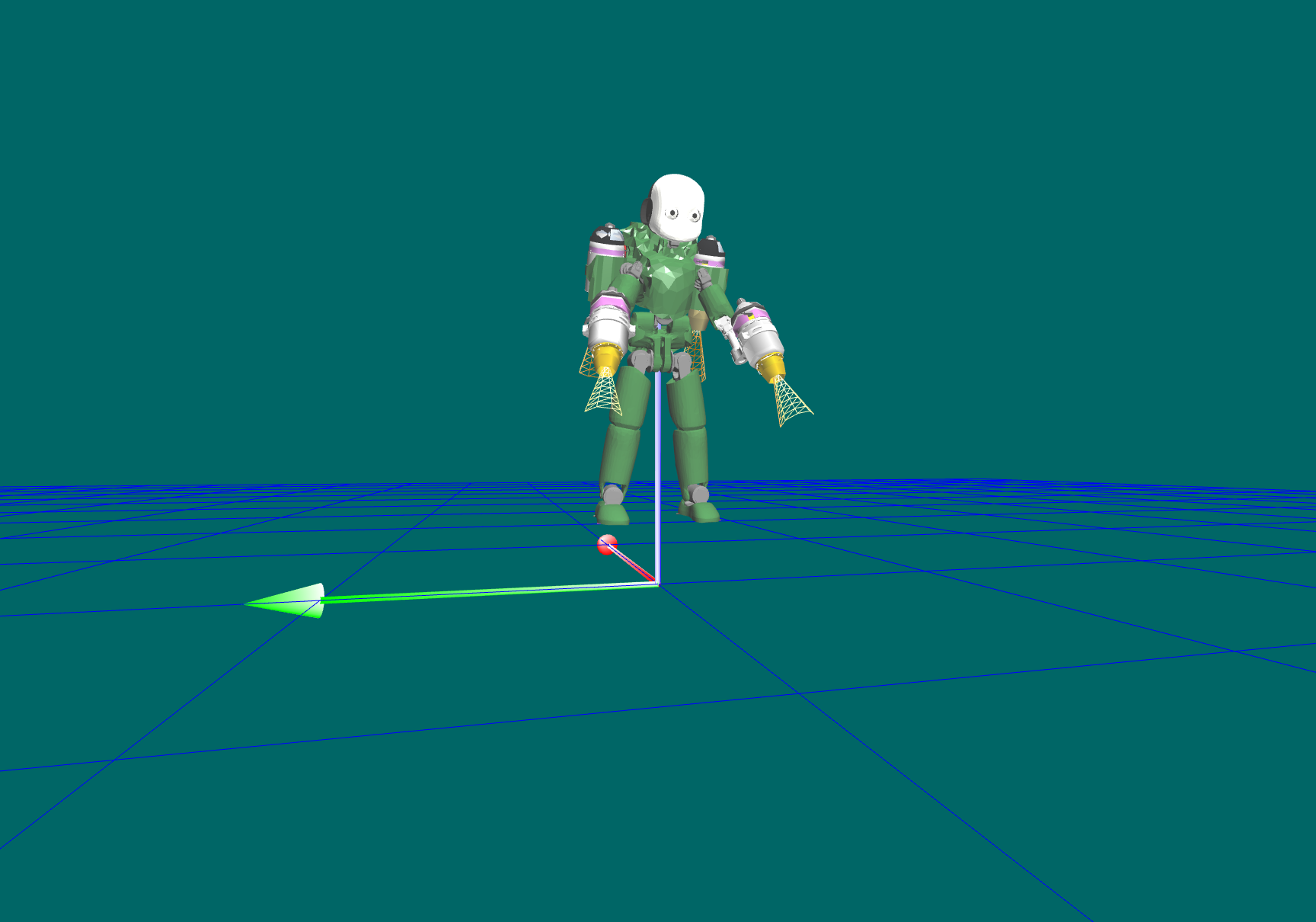}
     \includegraphics[width=0.33\textwidth]{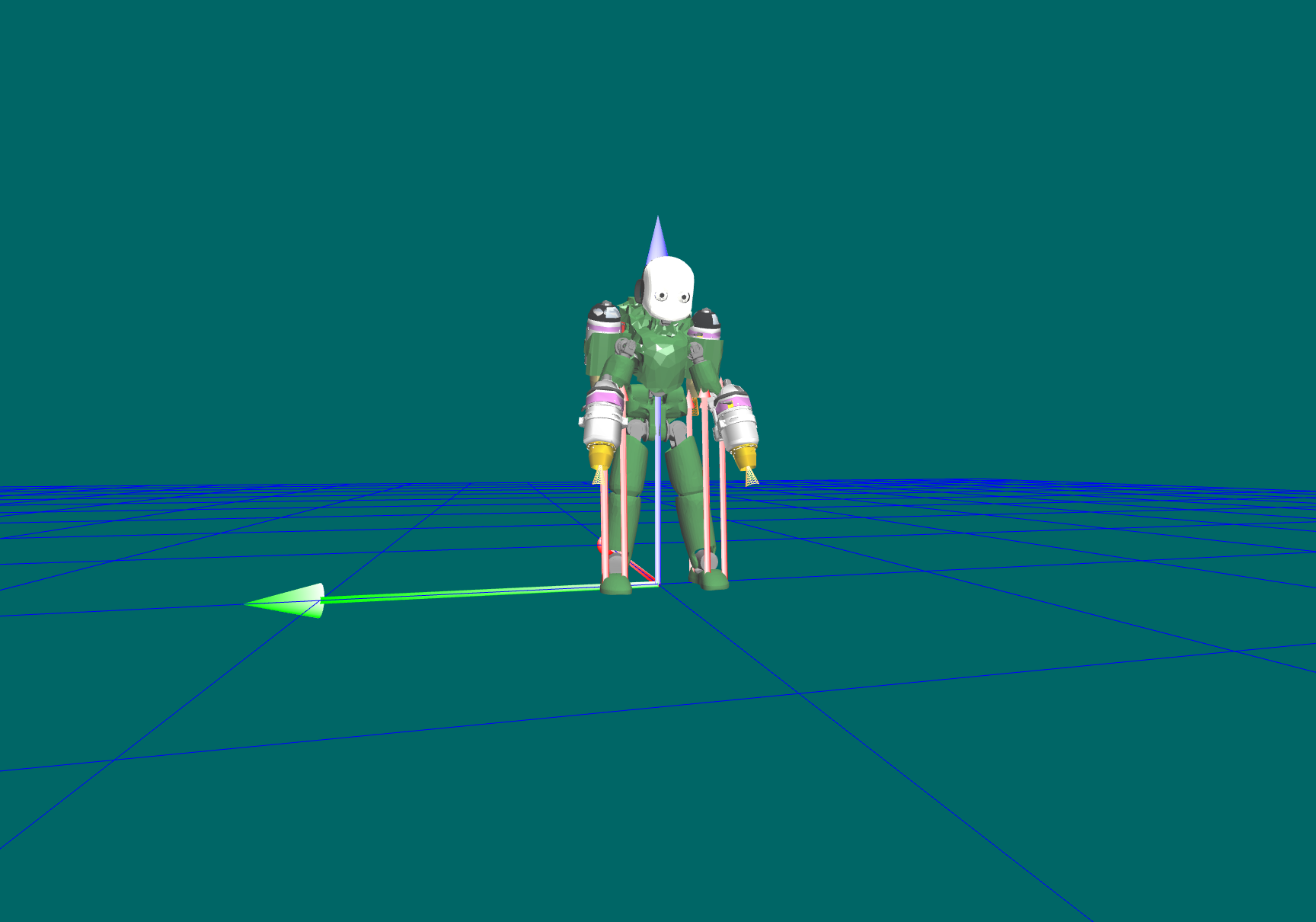}
    \end{minipage}} \par
    \subfloat[Trajectories of throttle, thrust and total vertical component of the contact forces during the take-off and landing. The red band represents the flight phase.\looseness=-1] {%
    \begin{minipage}{\linewidth} \label{fig:tol_plots}
    \includegraphics[width=0.33\textwidth]{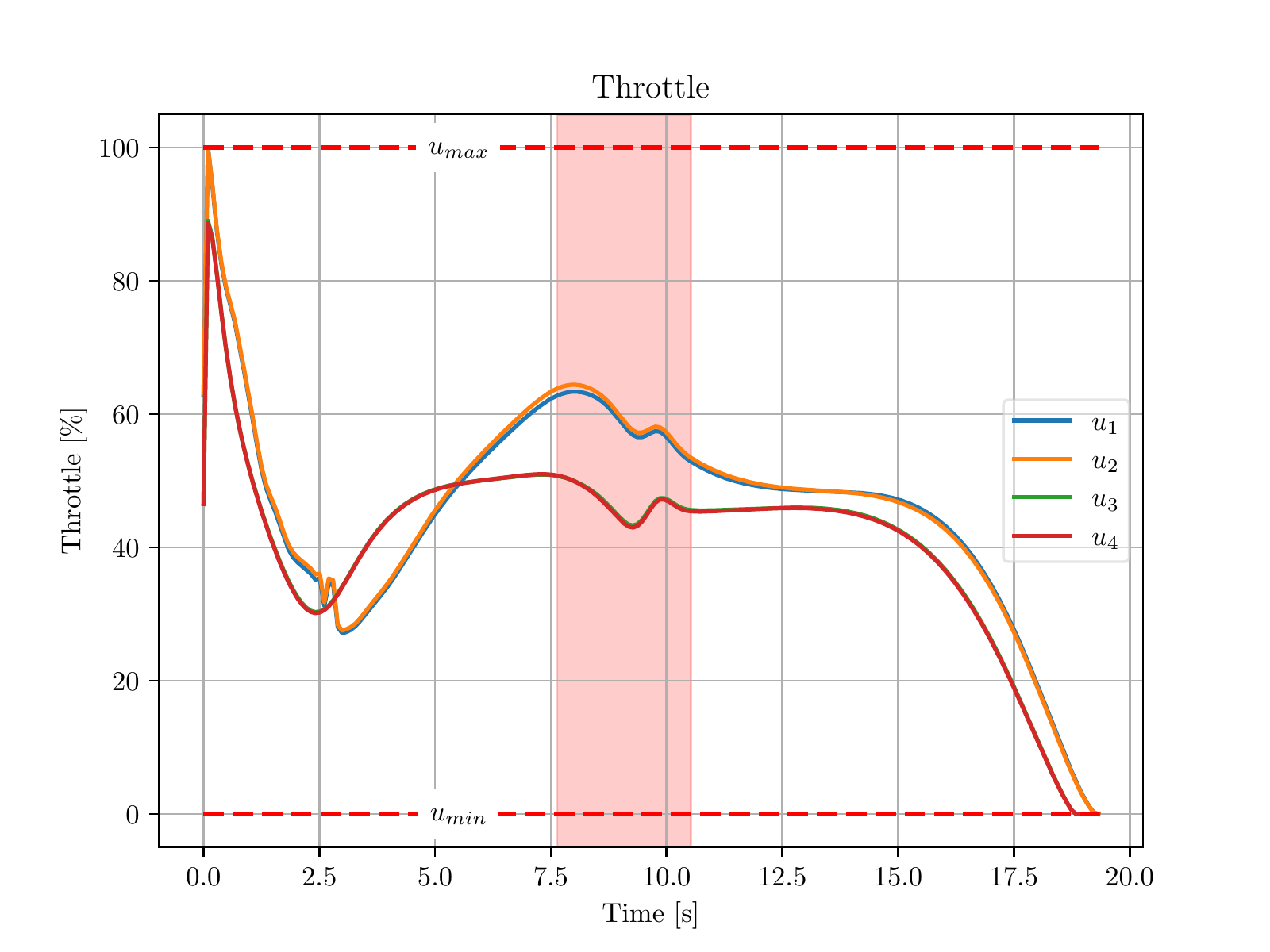}
     \includegraphics[width=0.33\textwidth]{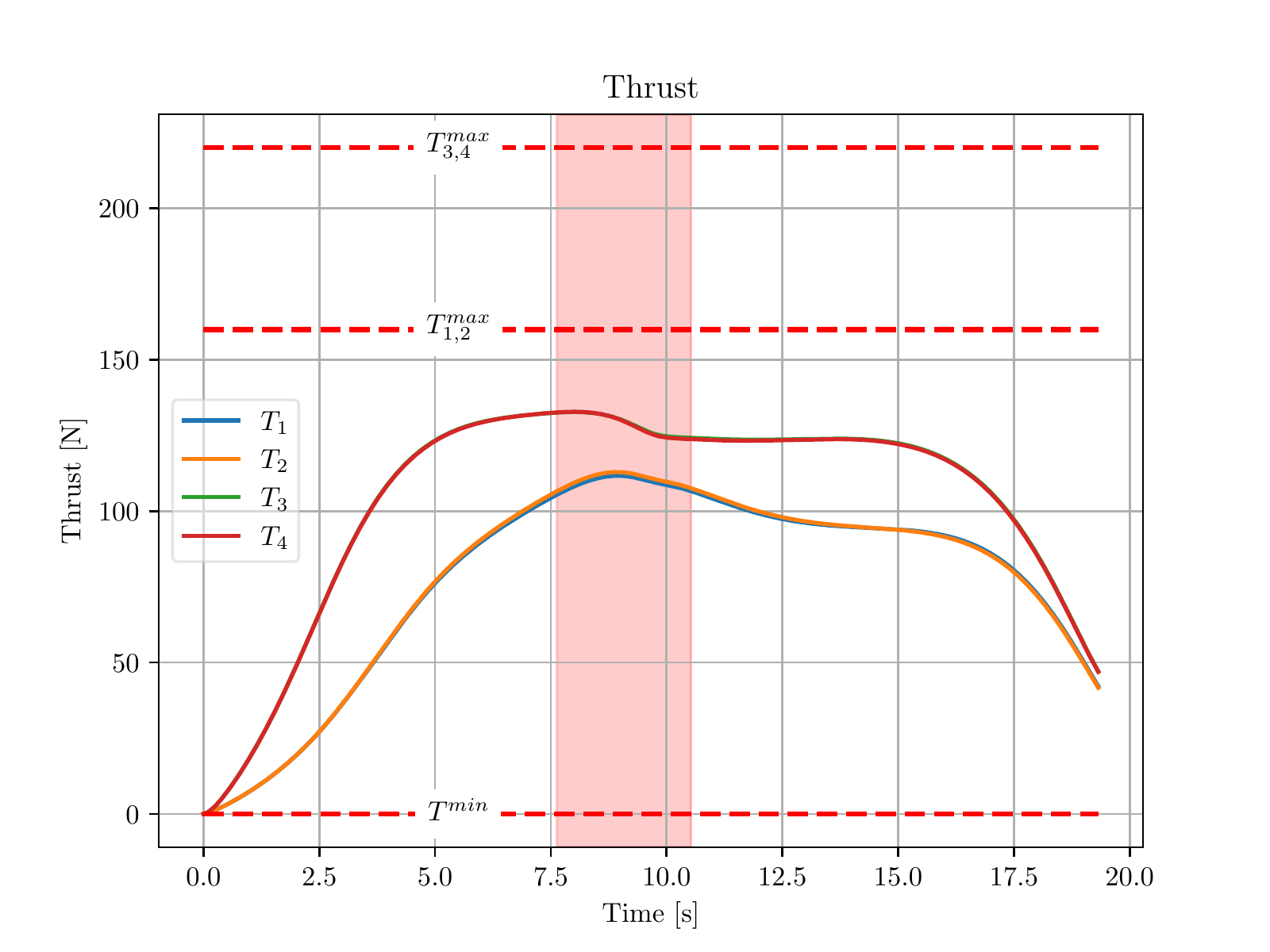}
     \includegraphics[width=0.33\textwidth]{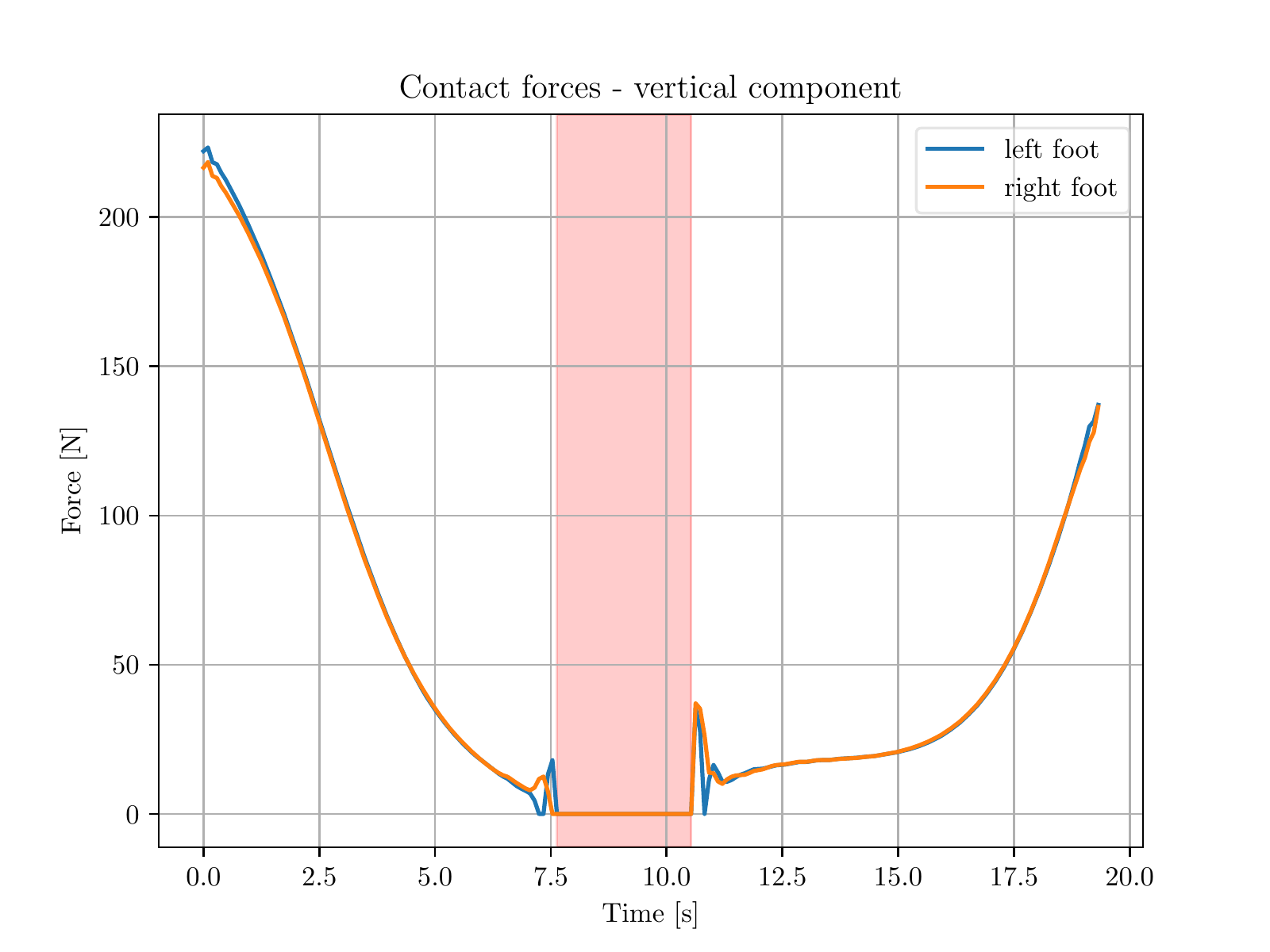}
     \end{minipage}}
  \end{myframe}
  \caption{Snapshots and trajectories of the robot during the take-off and landing.} 
\end{figure*}

Different from the scenarios above, the second context we test our approach is a vertical take-off and landing that shall be generated as a single instance of the optimisation problem. For this reason, we need to increase the number of knots of the optimisation horizon to 200. The initial and final CoM position is set to \SI{0.57}{m} while at the intermediate knot we request a CoM height higher than \SI{0.7}{m}. Snapshots of this manoeuvre are in Fig~\ref{fig:tol_snapshots}. The robot decreases the contact forces and increases the thrust. The trajectory of the CoM reaches the maximum height and smoothly comes back to the initial position. Note that the weights in the cost function need to be tuned properly, otherwise the trajectory would resemble more a jump than take-off and landing.\looseness=-1

\subsection{Walking to Flight transition}

\begin{figure*}[tbp]
    \centering
    \begin{myframe}{Walking to flight transition}
    \subfloat[Snapshots of the transition from  walking to flight] {%
    \begin{minipage}{\linewidth} \label{fig:transition_snapshots}
     \includegraphics[width=0.245\textwidth]{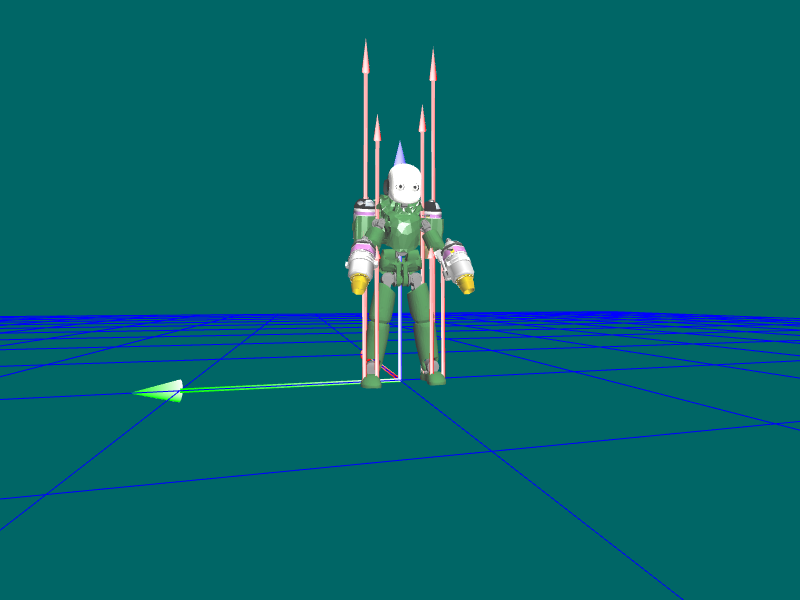}
     \includegraphics[width=0.245\textwidth]{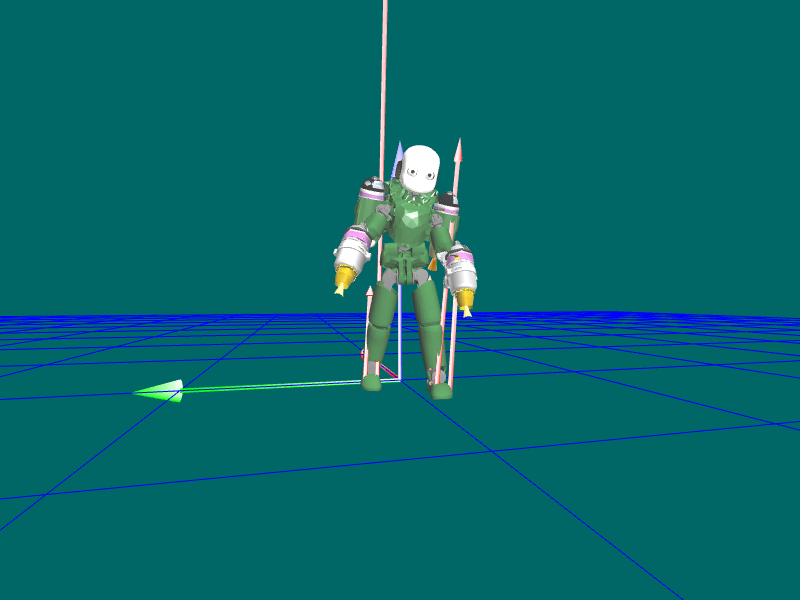}
     \includegraphics[width=0.245\textwidth]{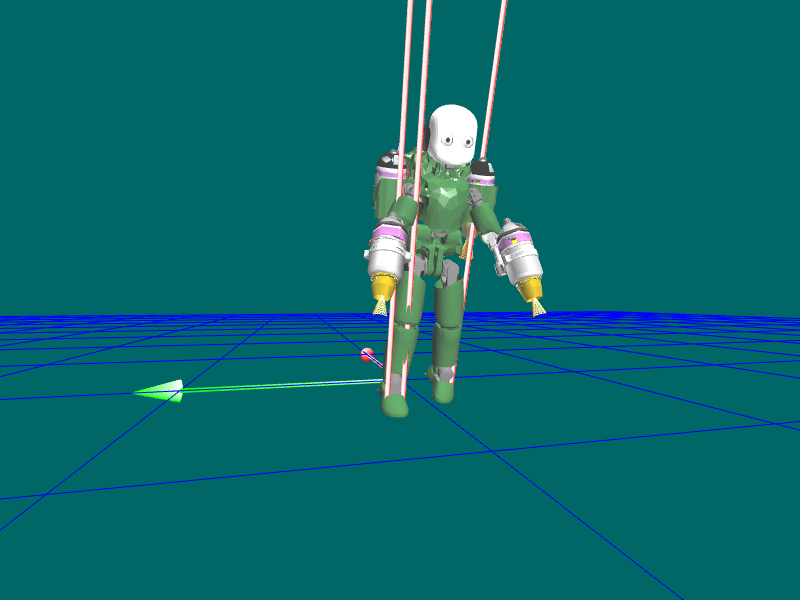}
     \includegraphics[width=0.245\textwidth]{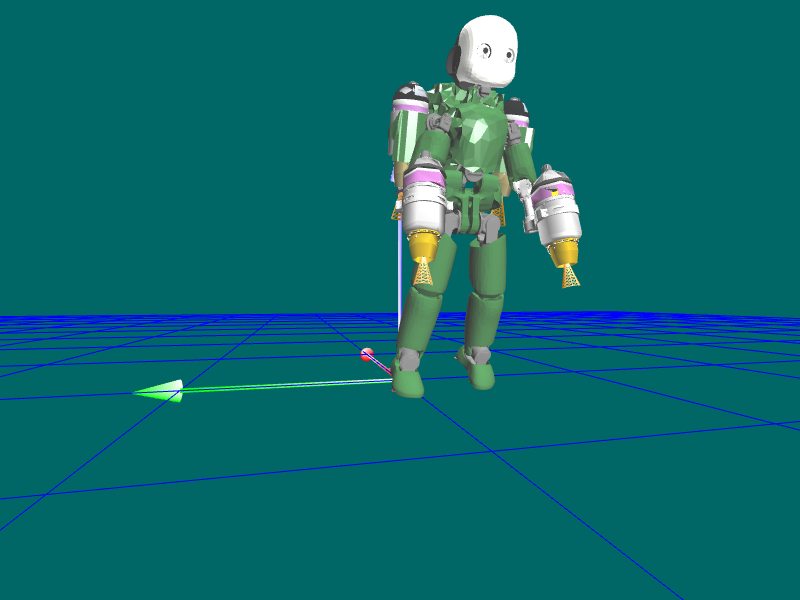}
    \end{minipage}} \par
    \subfloat[Trajectories of throttle, thrust and total vertical component of the contact forces during the transition from walking to flying. The red band represents the flight phase.] {%
    \begin{minipage}{\linewidth} \label{fig:transition_plots}
    \includegraphics[width=0.33\textwidth]{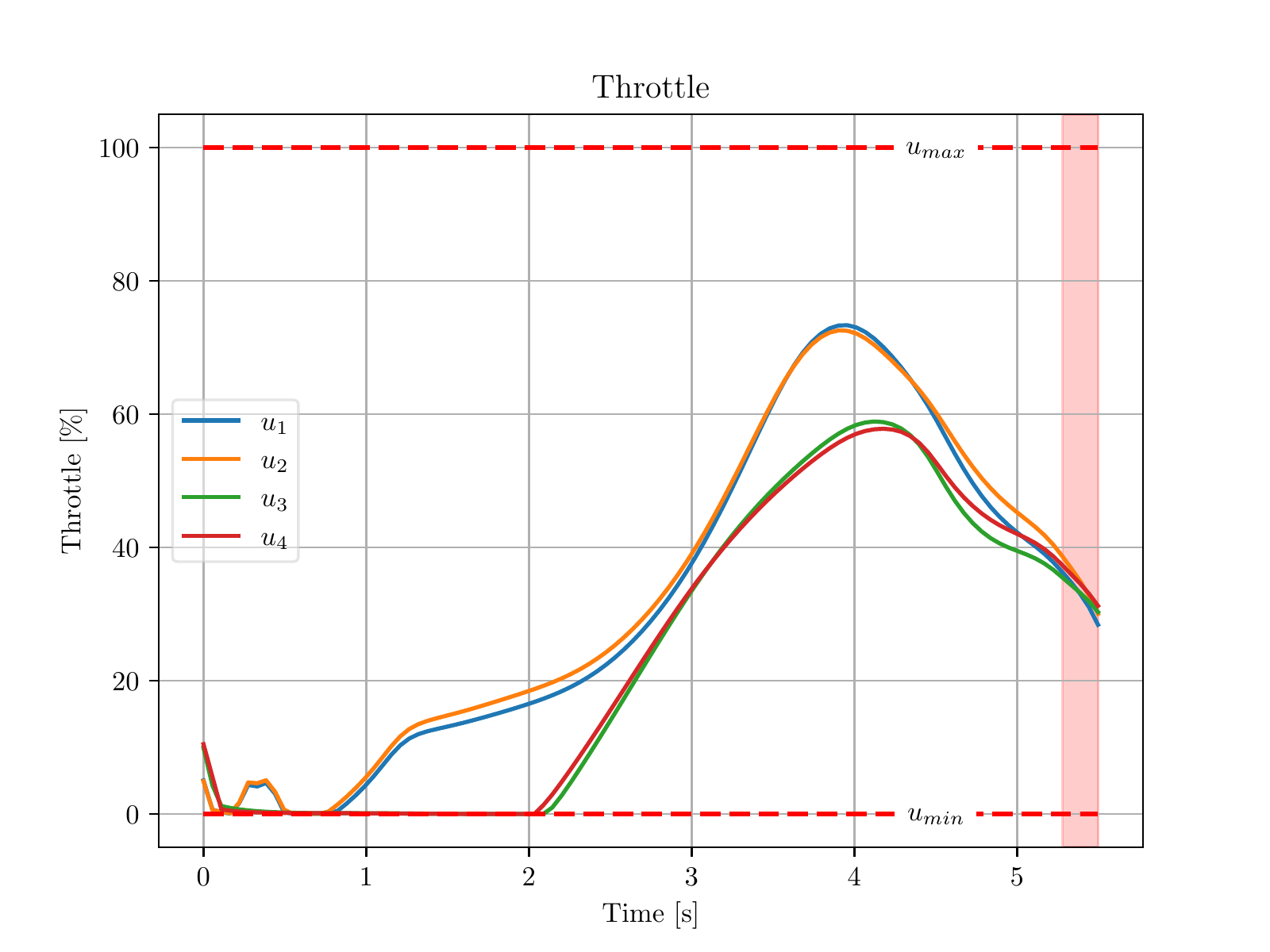}
     \includegraphics[width=0.33\textwidth]{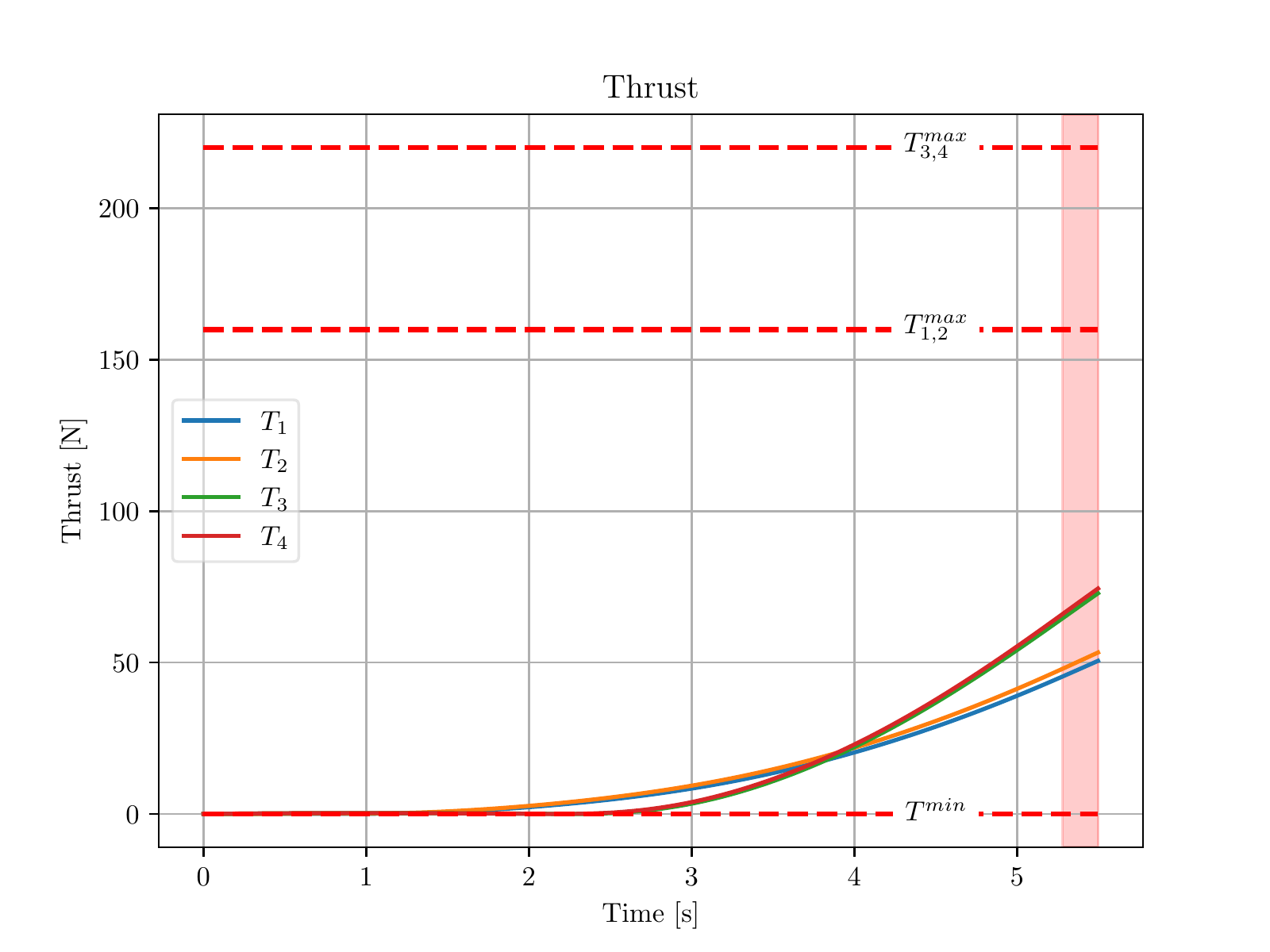}
     \includegraphics[width=0.33\textwidth]{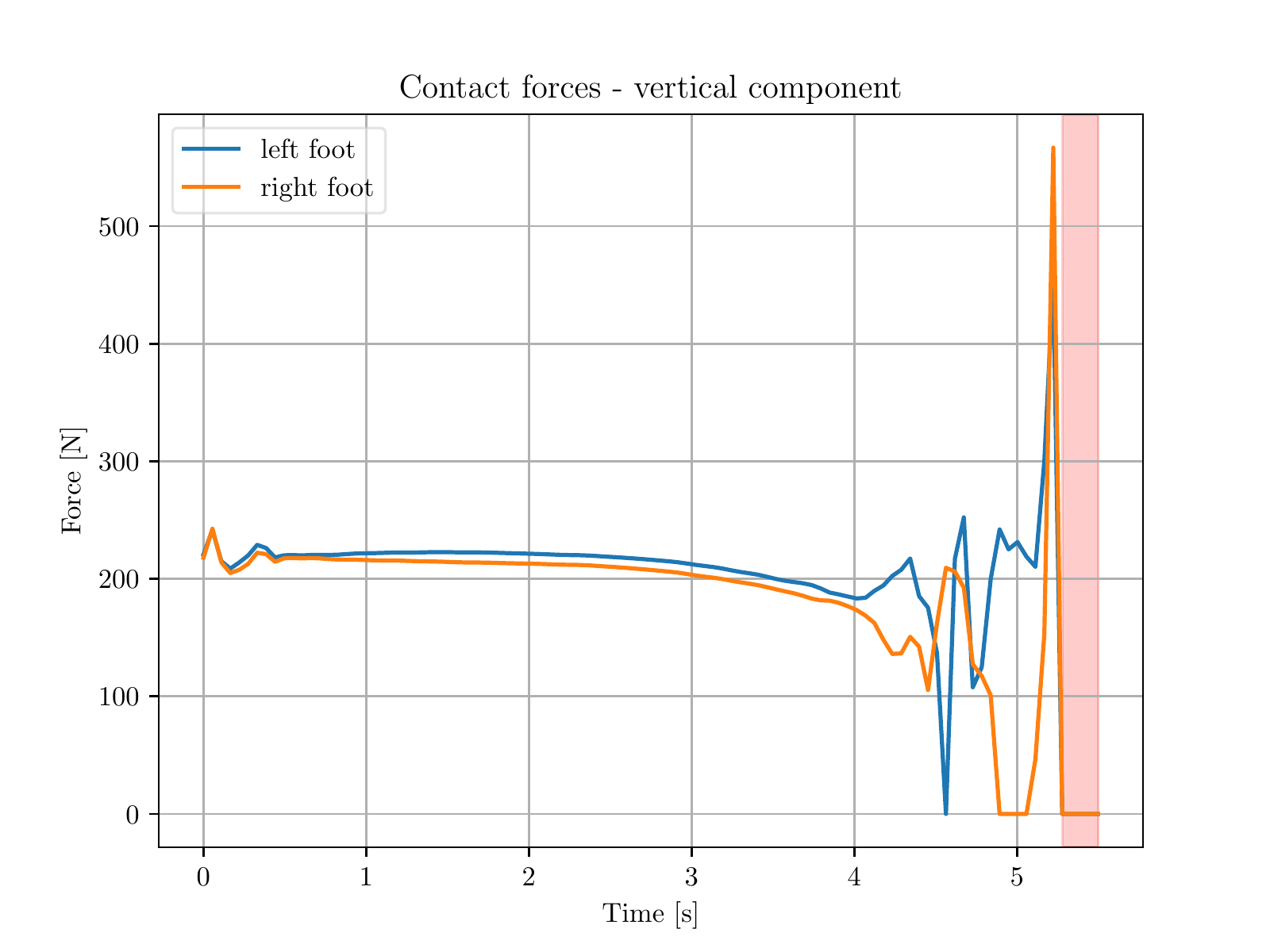}
     \end{minipage}}
  \end{myframe}
  \caption{Snapshots and trajectories of the transition from walking to flight.} 
\end{figure*}

We test our approach in a context in which two different locomotion patterns are mixed. The robot starts at an initial CoM position at \SI{0.57}{m} and transitions from walking to flying locomotion going \SI{0.9}{m} forward and reaching \SI{0.7}{m}. The number of knots of the optimisation horizon is equal to 100. Snapshots of this behaviour are in Fig~\ref{fig:transition_snapshots}. The robot moves a few steps forward while increasing the thrust. Eventually, the contacts break and the robot detaches from the ground. Note that the contact sequence in the legged locomotion phase is not predefined and emerges directly from the formulation. 

\subsection{Jumping}

To validate the approach for generating multimodal locomotion, the fourth scenario consists of the robot taking a jump. The number of optimisation knots is set to $N=60$. The initial and the final CoM position height equal $ \SI{0.57}{m}$. We add an additional constraint on the CoM height at the middle-knot $x[N/2] \ge \SI{0.75}{m}$. The CoM height decreases and then accelerates vertically until the contacts break. After the flight phase, the robot lands and returns to the initial configuration.\looseness=-1

\begin{figure}[tpb]
\vspace{0.3cm}
    \begin{minipage}{\linewidth}
    \centering
    \includegraphics[trim={0 5.5cm 0 0}, clip, width=.3\columnwidth]{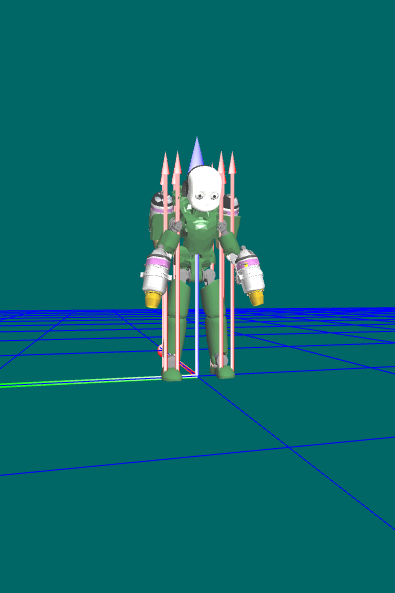} 
    \includegraphics[trim={0 5.5cm 0 0}, clip, width=.3\columnwidth]{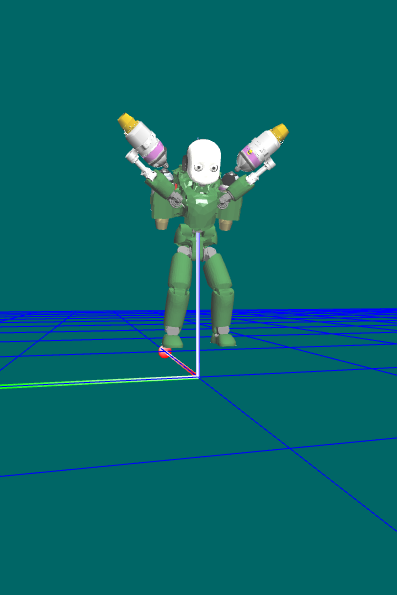} 
    \includegraphics[trim={0 5.5cm 0 0}, clip, width=.3\columnwidth]{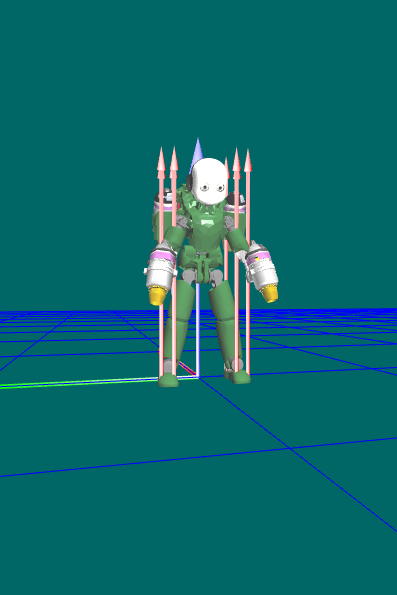}
    \end{minipage}
    \label{fig:jump}
    \caption{Snapshots of the robot jumping.}
    \vspace{-0.3cm}
\end{figure}

%% file: tex/conclusions.tex
\section{Conclusions}
\label{sec:conclusions} 

This work proposes an extension of the so-called \textit{kino-dynamic} planning to platforms that have multimodal locomotion capabilities, such as a flying humanoid robot. We formulate the trajectory planner as an optimisation problem, which optimises the contact sequence and robot motions, subject to the jet dynamics. Numeric examples with the model of iRonCub, our flying humanoid robot, show the effectiveness of the proposed approach. The trajectory optimisation can generate take-off and landing motion, as well as more general behaviours such as gait patterns or jumping. Large computational time makes our trajectory optimisation suitable for planning purposes rather than as a control strategy. Future work focuses on the design of a whole-body control that stabilizes the generated trajectories. \looseness=-1